\documentclass{article}
\usepackage{float}
\usepackage[margin=1.0in]{geometry}
\usepackage{abstract}
\usepackage{booktabs}
\usepackage{multirow}
\usepackage{graphicx}
\usepackage{wrapfig}
\usepackage{subfigure}
\usepackage{caption}
\usepackage[dvipsnames,svgnames, table,xcdraw]{xcolor}
\usepackage{relsize}
\usepackage{tcolorbox}
\usepackage{listings}
\usepackage{algorithm}
\usepackage{algorithmic}
\usepackage{comment}
\usepackage{graphicx}
\usepackage{booktabs}
\usepackage{multirow}
\usepackage{enumitem}
\usepackage{url}
\usepackage{mathpazo}
\usepackage{amsmath,amssymb,amsthm,amsfonts,bm}
\usepackage[toc,page,header]{appendix}
\usepackage{fancyhdr}
\usepackage[backend=biber,style=nature,natbib=true,maxbibnames=99,minalphanames=3]{biblatex}
\usepackage[colorlinks=true]{hyperref}
\usepackage[T1]{fontenc}
\usepackage{cleveref}
\usepackage{adjustbox}
\usepackage{wrapfig}

\usepackage{auth_detailed}

\counterwithout{figure}{section}
\counterwithout{table}{section}
\definecolor{darkgoldenrod}{rgb}{0.72, 0.53, 0.04}
\definecolor{backgroundcolor}{RGB}{250, 250, 252}   
\definecolor{keywordcolor}{RGB}{30, 0, 178}       
\definecolor{stringcolor}{RGB}{204, 0, 102}        
\definecolor{numbercolor}{RGB}{0, 128, 128}        
\definecolor{emphcolor}{RGB}{30, 0, 178}            
\definecolor{commentcolor}{RGB}{0, 128, 0}       
\definecolor{basiccodecolor}{RGB}{61, 61, 61}

\lstdefinestyle{customstyle}{
    backgroundcolor=\color{backgroundcolor},   
    commentstyle=\color{commentcolor},
    keywordstyle=\color{keywordcolor},
    numberstyle=\color{numbercolor},
    stringstyle=\color{stringcolor},
    basicstyle=\color{basiccodecolor}\ttfamily\footnotesize,
    breakatwhitespace=false,         
    breaklines=true,                 
    captionpos=b,                    
    keepspaces=true,                 
    numbers=left,     
    basicstyle=\color{basiccodecolor}\ttfamily\footnotesize,
    numbersep=5pt,             
    xleftmargin=2em,
    xrightmargin=2em,
    showspaces=false,                
    showstringspaces=false,
    showtabs=false,                  
    tabsize=1,
    frame=single,
    framesep=5pt,
    framexleftmargin=1.5em,
    framexrightmargin=1.5em,
    framextopmargin=1pt,
    framexbottommargin=1pt,
    aboveskip=10pt,
    belowskip=10pt,
    breaklines=true,
    breakautoindent=true,
    emph={textgrad, tg, Variable, MultipleChoiceTestTime,
    TextualGradientDescent, BlackboxLLM},             %
    emphstyle={\color{emphcolor}},
    extendedchars=true,
}

\definecolor{logocolor}{RGB}{30, 0, 178}

\definecolor{darkerlogocolor}{RGB}{20, 0, 145}  

\newtcolorbox{ttcolorbox}[1][]{colframe=darkerlogocolor, colback=darkerlogocolor!4!white, title=#1}

\newtcolorbox{apxtcolorbox}[1][]{colframe=black, colback=black!3!white, title=#1}

\definecolor{ForestGreen}{RGB}{34,139,34} 
\definecolor{RoyalBlue}{RGB}{65,105,225}

\newcommand{\ie}{\em{i.e.}}
\newcommand{\eg}{\em{e.g.}}

\usepackage{amsmath,amsfonts,bm}

\def\eqref#1{equation~\ref{#1}}

\def\1{\bm{1}}

\def\ve{{\bm{e}}}

\def\vv{{\bm{v}}}

\DeclareMathAlphabet{\mathsfit}{\encodingdefault}{\sfdefault}{m}{sl}
\SetMathAlphabet{\mathsfit}{bold}{\encodingdefault}{\sfdefault}{bx}{n}

\hypersetup{
  colorlinks=true,
  urlcolor=BrickRed,
  citecolor=RoyalBlue,
  linkcolor=BrickRed
} 
\BiblatexSplitbibDefernumbersWarningOff
\newcommand{\model}[1]{K-Flow}

\title{
Flow Along the $K$-Amplitude for Generative Modeling
}
\author{\name Weitao Du$^{1}$ \email duweitao.dwt@alibaba-inc.com\\
\name Jiasheng Tang$^{1}$ \email jiasheng.tjs@alibaba-inc.com\\
\name Shuning Chang$^{1}$ \email changshuning@alibaba-inc.com\\
\name Yu Rong$^{1}$ \email yu.rong@hotmail.com\\
\name Fan Wang$^{1}$ \email wangfan@alibaba-inc.com\\
\name Shengchao Liu$^{2}$ \email shengchao.liu@berkeley.edu\\
$^{1}$Alibaba DAMO Academy\\
$^{2}$University of California, Berkeley\\
\email \begin{normalsize}
\end{normalsize}
}

\begin{document}

\fancyhead[R]{\model{}}
\setlength{\headheight}{13pt}
\pagestyle{fancy}

\newcounter{suppfigure}
\newcounter{supptable}
\makeatletter
\newcommand\suppfigurename{Supplementary Figure}
\newcommand\supptablename{Supplementary Table}
\newcommand\suppfigureautorefname{\suppfigurename}
\newcommand\supptableautorefname{\supptablename}
\let\oldappendix\appendix
\renewcommand\appendix{%
    \oldappendix
    \setcounter{figure}{0}%
    \setcounter{table}{0}%
    \renewcommand\figurename{\suppfigurename}%
    \renewcommand\tablename{\supptablename}%
}
\makeatother

\maketitle

\renewcommand{\abstractnamefont}{\normalfont\normalsize\bfseries}
\renewcommand{\abstracttextfont}{\normalfont\normalsize}

\begin{figure}[h]
    \centering
    \includegraphics[width=\linewidth]{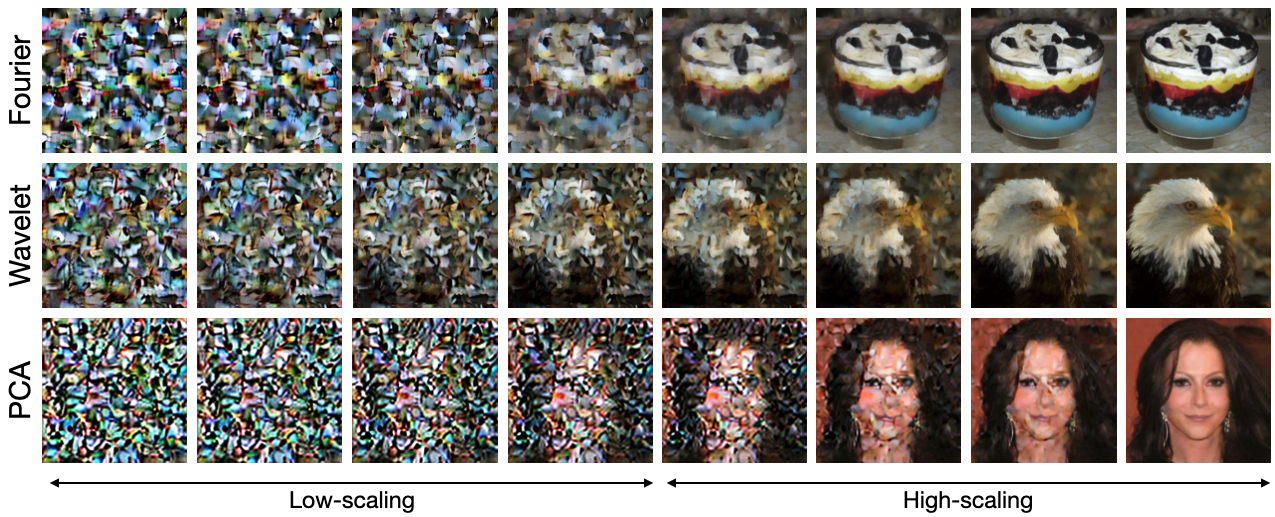}
    \vspace{-5ex}
    \caption{\small Unconditional generation using \model{} using three types of $K$-amplitude decomposition: Fourier, Wavelet, and PCA.
    }
    \label{fig:01_starting_figure}
\end{figure}

\renewcommand{\abstractname}{Abstract}
\begin{abstract}
\abstracttextfont
In this work, we propose a novel generative learning paradigm, K-Flow, an algorithm that flows along the $K$-amplitude. Here, $k$ is a scaling parameter that organizes frequency bands (or projected coefficients), and amplitude describes the norm of such projected coefficients. By incorporating the $K$-amplitude decomposition, K-Flow enables flow matching across the scaling parameter as time. We discuss three venues and six properties of K-Flow, from theoretical foundations, energy and temporal dynamics, and practical applications, respectively. Specifically, from the practical usage perspective, K-Flow allows steerable generation by controlling the information at different scales. To demonstrate the effectiveness of K-Flow, we conduct experiments on unconditional image generation, class-conditional image generation, and molecule assembly generation. Additionally, we conduct three ablation studies to demonstrate how K-Flow steers scaling parameter to effectively control the resolution of image generation.
\end{abstract}

\newpage
\tableofcontents
\newpage
\section{Introduction} \label{sec:introduction}

Generative Artificial Intelligence (GenAI) represents a pinnacle achievement in the recent wave of AI advancements. This field has evolved from foundational methods such as autoregressive models (AR)~\citep{radford2018improving}, energy-based models (EBMs)~\citep{hinton2002training,carreira2005contrastive,lecun2006tutorial,gutmann2010noise,song2021train}, variational auto-encoders (VAEs)~\citep{kingma2013auto}, and generative adversarial networks (GANs)~\citep{goodfellow2014generative}, to the most cutting-edge flow-matching (FM) framework~\citep{lipman2022flow,liu2022flow,albergo2022building}.

Among these, flow matching (FM) stands out as a density transport method that converts an initial simple distribution into a complex target distribution through continuous-time flow dynamics. For instance, in the context of image generation, FM learns to map a random Gaussian distribution to the pixel-space distribution of images. This process, termed continuous \textit{flow}, is governed by a localized \textit{k-dependent vector field} (or velocity field) and produces a \textit{time-dependent density path}, which represents the evolution of the probability distribution over time. As a versatile framework, FM can incorporate a diffusion density path, linking it to established methods such as denoising score matching (DSM)~\citep{vincent2011connection,song2019generative} and the denoising diffusion probabilistic model (DDPM)~\citep{ho2020denoising}.

\textbf{Motivation.}
Current generative models lack principled mechanisms for multi-scale control of synthesized content. While existing approaches enable coarse attribute editing, precise preservation of certain frequency structures while modifying others ({\eg}, high-frequency details) - a critical requirement for applications like general image restoration and specific data generation like medical and science data - remains an open challenge. This motivates our development of a frequency-domain grounded framework with an inherent scale hierarchy. From the pure generation capability side, current research like~\citep{skorokhodov2025improving} also suggests explicit frequency consistent loss as a general regularization loss for auto-encoders. Thus, we are interested in introducing frequency regularization along the generation path. Combined with frequency-aware autoencoders, this approach may open the door to a new paradigm for data generation.

\textbf{Key Concepts.}
We first introduce several core concepts. The \textit{scaling parameter $k$} can be interpreted as a measure to organize the frequency bands (or coefficients) of physical objects or processes~\citep{cardy1996scaling,luijten1996finite,behan2017scaling,bighin2024universal}\footnote{Notably, we use ``scaling parameters'' when discussing parameterization, and ``scale'' in other contexts.}, and \textit{amplitude} refers to the norm of coefficients obtained after projecting data concerning the scaling parameter $k$, which we term the \textit{$K$-amplitude space}, or equivalently, \textit{scaling-amplitude space}. The underlying intuition behind the utility of $K$-amplitude space is that multi-scaling modeling inherently aligns more naturally with data structures in the $K$-amplitude space, {\ie}, lower $k$ tend to have higher amplitudes, as observed in multi-resolution image modeling~\citep{Abry1995}.

\textbf{Our Method.}
Such an understanding of scaling parameter and $K$-amplitude space inspires a new paradigm for generative modeling, which we term \textbf{K Flow Matching (\model{})}. In essence, \model{} performs flow along the $K$-amplitude. There are two main components in \model{}, and the first is the $K$-amplitude decomposition. The $K$-amplitude decomposition encompasses a family of transformations, and in this work, we explore three types: Wavelet, Fourier, and principal component analysis (PCA) decomposition, as illustrated in~\Cref{fig:01_starting_figure}. Building on this, the second component in \model{} is its flow process. \model{} applies a $K$-amplitude transformation to project data from the spatial space into the $K$-amplitude space, learns a time-dependent velocity field in this space accordingly, and subsequently maps it back to the spatial space for velocity matching. A detailed pipeline is provided in~\Cref{fig:pipeline_figure}. Next, we will discuss the strengths of \model{} through six properties, which can be organized into three categories: theoretical foundations (properties a \& b), energy and temporal dynamics (properties c \& d), and practical applications (properties e \& f).

\textbf{Our Results.}
We conduct experiments on image and molecule generation to verify the effectiveness of \model{}. Quantitatively, \model{} achieves competitive performance in unconditional and class-conditional image generation and molecular assembly. Qualitatively, we conduct three ablation studies to demonstrate the steerability of \model{}: controllable class-conditional generation and scaling-controllable generation.

\subsection{Properties of \model{}}
(a) \model{} provides \textbf{a first-principle way to organize the scaling $k$}. 
Unlike perception-based computer vision tasks, which often favor certain scaling (frequency) bands, a $K$-amplitude based generative model strives for an optimal organization of all scalings to ensure that the final generated sample is of high fidelity. By constructing $K$-amplitude scaling-based vector fields, the integrated flow naturally incorporates all scaling information, and the conditional flow matching training objective provides a perfect trade-off of accuracy-efficiency inside localized scalings. We will also demonstrate how different discretizations of \model{} with related works, highlighting the connections and integrations with existing methods in the field. 

(b) \model{} enables \textbf{multi-scale modeling in the $K$-amplitude space}.
Compared to the original data space, such as the pixel space in images, the $K$-amplitude space provides a more natural perspective for defining and analyzing multi-scale information, namely, $K$-amplitude decomposition empowers \model{} for effective multi-scale modeling. By decomposing the feature representation into multiple scaling components in the $K$-amplitude space, \model{} associates each scaling with an amplitude. Higher values of $K$-amplitude correspond to higher-frequency information, capturing fine-grained details, while lower values encode lower-frequency information, representing more coarse-grained features. Let us take the image for illustration. Images inherently exhibit a hierarchical structure, with information distributed across various resolution levels. Low-resolution components capture global shapes and background information, while high-resolution components encode fine details like textures, often sparse and localized. By projecting these components into the $K$-amplitude space, \model{} captures such hierarchical information effectively and naturally, enabling precise modeling of the interplay between scales.

(c) \model{} supports \textbf{a well-defined scale along with energy}.
The amplitude is also used to reflect the \textit{energy} level at each scale of the data. In physics, it is proportional to the square of the amplitude. In comparison, for the modeling on the original data space, though we can inject application-specific inductive bias, such as multiple pixel resolutions for images, they do not possess a natural energy concept.

(d) \model{} interprets \textbf{scaling as time}.
From elucidating the design space of the traditional flow matching perspective, \model{} re-defines the artificial time variable (or the signal-to-noise ratio variable proposed in \citep{kingma2021variational}) as the ordering index of frequency space. In this context, the artificial time variable effectively controls the traversal through different levels of a general notion of frequency decompositions, scaling each frequency component appropriately. This perspective aligns with the concept of renormalization in physical systems, where behavior across scales is systematically related.

(e) \model{} supports the \textbf{fusion of intra-scaling and inter-scaling modeling.}
\model{} flows across scaling as time, and namely, \model{} naturally merges the intra- and inter-scaling during the flow process. Thus the key module turns to the smooth interpolant, as will be introduced in~\Cref{sec:method}. This is in comparison with existing works on multi-modal modeling~\citep{burt1987laplacian,tian2024visual,atzmon2024edify}, where the special design of the intra-scaling and inter-scaling is required.

(f) \model{} supports \textbf{explicit steerability}.
The flow process across scales enables \model{} to control the information learned at various hierarchical levels. This, in turn, allows finer-grained control of the generative modeling, facilitating more precise and customizable outputs. By understanding and leveraging \model{}'s steerability, its utility can be significantly enhanced across diverse domains, including Artificial Intelligence-Generated Content (AIGC), AI-driven scientific discovery, and the safe, responsible development of AI technologies.
\section{Background} \label{sec:background}

\subsection{Scaling Parameter $k$, Amplitude, and $K$-amplitude Decomposition}

\begin{figure}[t]
    \centering
    \includegraphics[width=\linewidth]{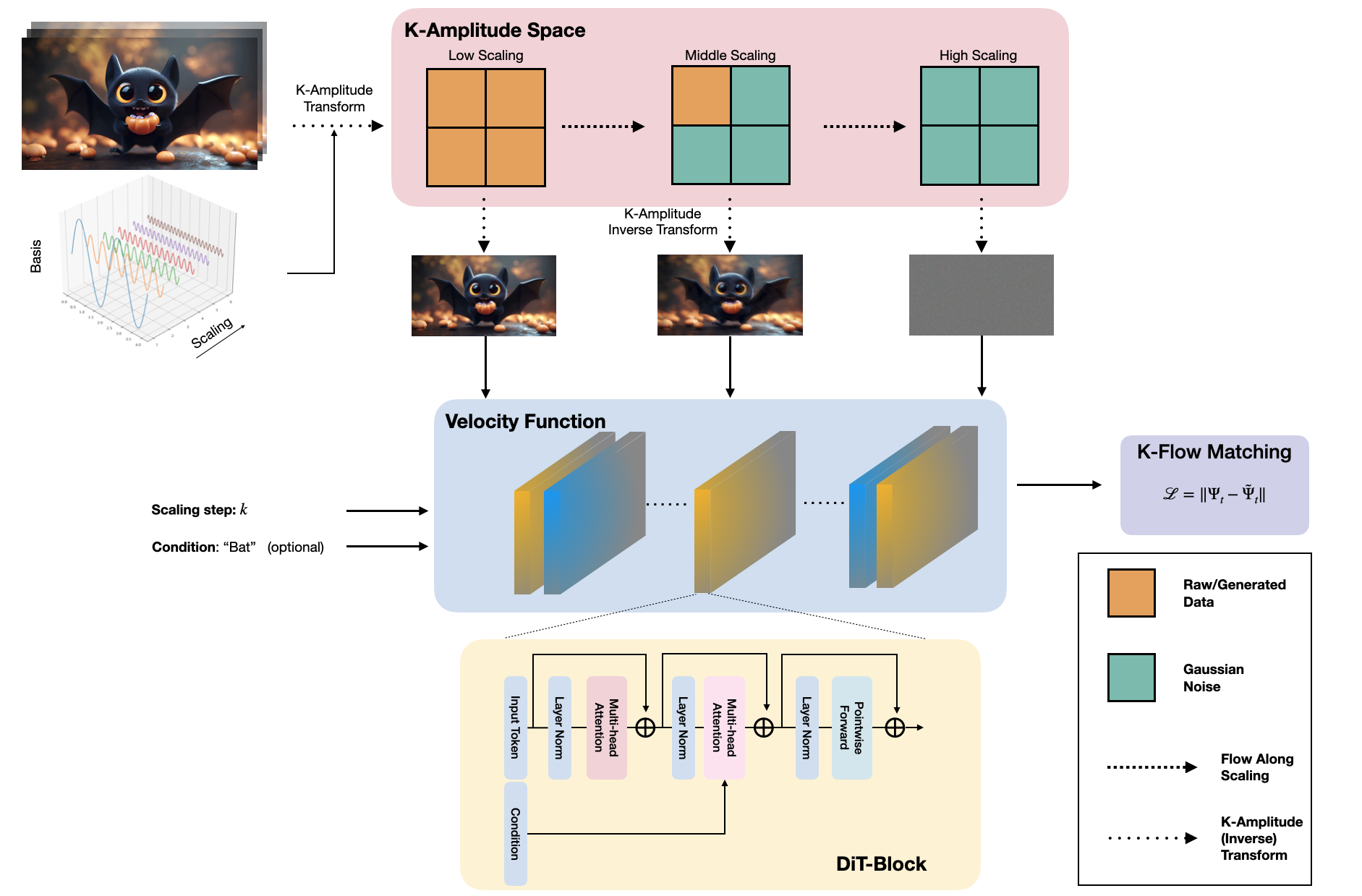}
    \vspace{-3ex}
    \caption{\small Pipeline of \model{}. In this figure, we have a bat figure as the input and three inverted images after three transformations at different granularities.}
    \label{fig:pipeline_figure}
\end{figure}

Our data generation framework leverages the implicit hierarchical structure of the data manifold. By `implicit', we refer to the hierarchical characteristics that emerge when a generalized $K$-amplitude decomposition is applied, transitioning the representation from the original data space to the $K$-amplitude space. Illustrations are in~\Cref{fig:pipeline_figure}.

More formally, we represent data as a signal $\phi: \mathbb{R}^d \rightarrow \mathbb{R}^m$, or a finite discretization of $\mathbb{R}^d$ and $\mathbb{R}^m$, where this signal function is equivalent to a vector. For example, in the case of image data, each pixel can be viewed as a signal mapping from x-y-RGB coordinates to a pixel intensity value, {\ie}, $\mathbb{R}^3 \rightarrow \mathbb{R}^1$. An alternative approach is to consider data as a high-dimensional vector $\mathbb{R}^{d \times m}$. However, treating data as signal functions provides a more natural fit for the decomposition framework introduced in this work.

Without loss of generality, we take $m=1$ for illustration. A $K$-amplitude decomposition involves the decomposition of a function using a complete basis set $\{\ve_j\}_{j=1}^n$, where $n$ can be infinite. We introduce a scaling parameter $k$, which partitions the set $\{\ve_i\}_{i=1}^n$ into subsets: $\{\ve_i\}_{i=1}^n = \bigcup_{k} \{e_k\}$, each with $n_k$ basis. Hence, signal $\phi$ is expressed as:
\begin{equation} \label{eq:signal_to_k_ampltitude}
\phi = \sum_k \phi_k,
\end{equation}
where $\phi_k := \sum_{j=1}^{n_k} (\phi \cdot \ve_{jk}) \ve_{jk}$ for $\ve_{jk} \in \{e_k\}$. Inspired by the concept of frequency amplitude, we also refer to the norm of $\phi_k$ as the $K$-amplitude. It is important to note that $k$ is termed the scaling parameter because it implies that a well-structured decomposition should ensure that the amplitude decays with increasing $k$~\citep{Field:87}.

We define $K$-amplitude decomposition (or equivalently, $K$-amplitude transform) $\mathcal{F}$ as the map that sends $\phi$ to the collection of $\phi_k$, and denote the collection of all $\{(\phi \cdot \ve_{jk})\ve_{jk}\}_j$ as $\mathcal{F}\{\phi\}(k)$. Then,
\begin{equation}
    \mathcal{F}\{\phi\} := \bigcup_{k} \mathcal{F}\{\phi\}(k).
\end{equation}
We further assume that this transform has an inverse, denoted by $\mathcal{F}^{-1}$.

\textbf{Splitting Probability.}
Denote the probability of data as $p_{\text{data}}$, then the transformations $\mathcal{F}$ and $\mathcal{F}^{-1}$ induce a probability measure on the associated $K$-amplitude space. In particular,  we denote the induced splitting probability of $\phi_k$ as $p(k)$ for each scaling parameter $k$.

In this work, we explore three types of $K$-amplitude decomposition: Wavelet, Fourier, and principal component analysis (PCA). In \Cref{sec:fourier_amplitude_decomposition_example}, we will provide a classic example using the Fourier frequency decomposition on the three-dimensional space. This example serves to illustrate the construction of the scaling parameter $k$ and $K$-amplitude.

\subsection{Example: Fourier Amplitude Decomposition} \label{sec:fourier_amplitude_decomposition_example}
Suppose the data $\phi: \mathbb{R}^3 \rightarrow \mathbb{R}$, is drawn from a certain function distribution $p_{\text{data}}$. The challenge of directly fitting the distribution $p_{\text{data}}$ is often complex and computationally demanding. Fourier frequency decomposition, however, offers a powerful technique to address this challenge by transforming $\phi$ into the Fourier space or Fourier domain. In what follows, we will use the terms `space' and `domain' interchangeably.

By applying Fourier frequency decomposition, we express $\phi$ as a sum of its frequency components. This transformation can potentially unveil the hidden structure within the distribution $p_{data}$, which is not apparent in the spatial or time domain, and it is thus beneficial for understanding the underlying patterns in the data manifold. To illustrate, the continuous Fourier transform $\mathcal{F}$ of data $\phi(x,y,z) : \mathbb{R}^3 \rightarrow \mathbb{R}$ is expressed as:
\begin{equation}
\mathcal{F}\{\phi\}(k_x, k_y, k_z) = \int_{-\infty}^{\infty} \int_{-\infty}^{\infty} \int_{-\infty}^{\infty} \phi(x, y, z) \, e^{-2\pi i (k_x x + k_y y + k_z z)} \, dx \, dy \, dz.
\end{equation}
After this transformation, the spatial variables \((x, y, z)\) are converted into frequency variables \((k_x, k_y, k_z)\), thereby representing the data in the frequency domain.

Note that the Fourier frequency is characterized by the high-dimensional vector representation $(k_x, k_y, k_z)$. For our purposes, we aim to distill the notion of frequency into a one-dimensional scaling parameter.
Namely, we define the scaling parameter $k$ as the diameter of the expanding ball in Fourier space: $k = \sqrt{k_x^2 + k_y^2 + k_z^2}$. This definition of $k$ provides a simple index that captures the overall scaling parameter of the frequency components in all directions. Moreover, we can decompose the Fourier transform $\mathcal{F}\{\phi\}$ into groups indexed by the scaling parameter $k$:
\begin{equation} \label{eq:Fourier_decomposition}
\mathcal{F}\{\phi\}(k) = \bigcup_{\sqrt{k_x^2 + k_y^2 + k_z^2} = k} \mathcal{F}\{\phi\}(k_x, k_y, k_z).
\end{equation}
Intuitively, $\mathcal{F}\{\phi\}(k)$ represents the set of all frequency components that share the same scaling parameter $k$. This grouping allows us to examine the contributions of various spatial frequencies of $\phi$ when viewed through the lens of frequency $k$. Furthermore, $\phi_k$ is just the summation of $\mathcal{F}\{\phi\}(k)$.

On the other hand, we can recover $\phi$ from $\mathcal{F}\{\phi\}$, because the Fourier transform is an invertible operation: $\phi = \mathcal{F}^{-1} \mathcal{F}\{\phi\}$. Such an invertibility establishes the Fourier transform as a valid example of $K$-amplitude decomposition. For discrete data, which inherently possess one highest resolution, the variables \((k_x, k_y, k_z)\) are situated on a discrete lattice rather than spanning the entire continuous space. Consequently, the scaling parameter $k$, derived from these discrete components, is itself discrete and bounded.

\subsection{Flow Matching}
In this work, we primarily focus on the flow matching (FM) generative models and their families~\citep{lipman2022flow,liu2022flow,albergo2022building}. In FM, the flow $\Psi_t$ is defined by solutions of an ordinary differential equation (ODE) system with a time-dependent vector field $\vv$:
\begin{equation} \label{eq: odeflow}
    \frac{d}{dt} \Psi_t(x) = \vv_t (\Psi_t (x)),
\end{equation}
and we focus on the probability transport aspects of $\Psi_t$. In particular, the flow provides a means of interpolating between probability densities within the sample space. Suppose $\Psi_t$ follows an initial probability $p_0$, then for $t>0$, $\Psi_t$ induces a probability measure $p_t$: $p_t(B) = p_0(\Psi_t^{-1}(B))$, where $B$ is a measurable set. Assume that $\Psi_t$ is differentiable, and define a surrogate velocity at time $t$ as $v_t(x_t, \theta)$ using a deep neural network with parameter $\theta$. Then the vector field matching loss is defined as:
\begin{equation}
\mathcal{L}_{\text{FM}} := \int \int_0^1 \, dx_0 \, dt \, \left\| \frac{d \Psi_t}{dt}(x_t) - v_t(x_t, \theta) \right\|^2.
\end{equation}
By aligning the learned vector field with the true gradient field of the frequency decomposition, this loss function ensures robust approximation and reconstruction of the data. Additionally, every interpolation $\pi(x_0, x_1)$ with a time-continuous interpolating function $f_t(x_0, x_1)$ between probabilities $p_0$ and $p_1$ induces a vector field $v_t$ through the continuity equation:
\begin{equation} \label{eq: abs match}
\frac{\partial p_t(x_t)}{\partial t} = - \nabla_x \left( p_t(x_t) v_t(x_t) \right),
\end{equation}
and $v_t$ is explicitly expressed as: $v_t = \frac{1}{p_t} \, \mathbb{E}_{\pi(x_0, x_1)}[\frac{\partial f_t(x_0, x_1)}{\partial t}]$. Although explicit matching of $v_t$ via the continuity equation is intractable, flow matching permits a conditional version:
\begin{align} \label{eq: conditional fm}
\mathcal{L}_{\text{CFM}} =  \mathbb{E}_{\pi(x_0, x_1)} &\int_0^t \, dt \, \left\| \frac{\partial f_t(x_0, x_1)}{\partial t} - v_t(x_t, \theta) \right\|^2 + \text{constant}.
\end{align}
As detailed in \Cref{sec:method}, our framework reinterprets the time variable $t$ as scaling parameter $k$. Our goal is to construct a $K$-amplitude-respected $\pi(x_0, x_1)$ with differentiable functions $f_k$.

\section{Methodology: \model{}} \label{sec:method}
In this section, we introduce \model{}. It is constructed from the collection of $\mathcal{F}\{\phi\}(k)$, indexed by the scaling parameter $k$. As we will demonstrate in~\Cref{sec:interpolants}, our approach is independent of the specific construction of the invertible transformation $\mathcal{F}$ and the explicit definition of $k$. This flexibility enables us to extend to various $K$-amplitude decompositions.

\subsection{$K$-amplitude Interpolants} \label{sec:interpolants}
According to the concept of stochastic interpolants~\citep{albergo2023stochastic}, all flow models can be viewed as constructing stochastic paths that interpolate between a known tractable prior distribution and an unknown target distribution, including flow matching~\citep{lipman2022flow}, rectified flow~\citep{liu2022flow}, and denoising diffusion~\citep{ho2020denoising}. By incorporating the scaling parameter $k$ for $K$-amplitude decompositions, we can formulate a stochastic interpolant that gradually emerges each amplitude component from white noise. Given that $k$ traverses monotonically from zero to a maximum value $k_{\text{max}}$, this process draws a natural analogy to continuous normalizing flows. Since we require $\mathcal{F}$ to be invertible, we can reconstruct the data once the complete spectrum in the $K$-amplitude space is generated.\looseness=-1

To build a \textbf{continuous flow} $\Psi_k$ out of \Cref{eq:signal_to_k_ampltitude}, we explore two paradigms in designing the interpolants: (1) We generalize the original discrete-valued $k$ to continuous values; (2) We ensure that the generation flow, which maps the white noise to the real data, remains invertible such that no information is lost throughout the process. Still taking the three-dimensional signal $\phi(x,y,z)$ and the Fourier transform $\mathcal{F}\{\phi\}$ as an example, we realize the second ingredient by introducing noise padding $\epsilon$ for each $k$ and define the \textbf{discrete flow} $\varphi_k$ as follows:
\begin{equation} \label{eq:K_amplitude_flow}
\varphi_k = \mathcal{F}^{-1} \left( \mathbb{I}_{\sqrt{k_x^2 + k_y^2 + k_z^2} \leq k} \cdot \mathcal{F}\{\phi\}(k_x, k_y, k_z) + \left(1 - \mathbb{I}_{\sqrt{k_x^2 + k_y^2 + k_z^2} \leq k} \right) \cdot \epsilon \right),
\end{equation}
where $\mathbb{I}$ is the indicator function that selects $K$-amplitude components up to the scaling step $k$. This formulation ensures that for each step $k$, the reconstruction incorporates the relevant \model{} components of data $\phi$ and pads the rest with noise $\epsilon$. Here, the noise $\epsilon$ is independently drawn from a known distribution ({\eg}, uniform or Gaussian) for each coordinate $(k_x, k_y, k_z)$. Through this construction, $\phi_k$ serves as a stochastic interpolant for the data $\phi$, ensuring that: $\lim_{k \rightarrow k_{\text{max}}} \varphi_k = \phi$, where $ k_{\text{max}} $ represents the maximum scaling parameter of data. This limit condition guarantees that as $k$ approaches its maximum value, the reconstructed $\varphi_k$ converges to the original data $\phi$. This behavior is pivotal for the accuracy and fidelity of the generative process. Conversely, $\phi_0$ simply follows the law of a tractable distribution.

\textbf{Inter-scaling Interpolant.}
Since most of the data we aim to generate is discrete in nature, the $(k_x,k_y,k_z)$ values in the $K$-amplitude decomposition are inherently defined on a lattice. Consequently, the derived scaling parameter $k$ also takes discrete values. This discreteness implies that $\varphi_k$ is originally defined only for discrete values of $k$. However, this discrete flow imposes a limitation: we cannot leverage the powerful flow-matching objective as the optimization framework, which requires taking derivatives with respect to continuous scaling step $k$.

To handle this issue, a straightforward approach is to extend $\varphi_k$ to continuous $k$ by intra-scaling interpolation. That is, we want a continuous flow $\Psi_k$, where $k \in [0, K]$ and satisfy $\Psi_k = \varphi_k$ for integer values of $k$. Let $t := k - \lfloor k \rfloor$ represent the continuous scaling step, where $\lfloor k \rfloor$ denotes the integer part of $k$. Then, the differentiable interpolation of $\Psi_{k}$ is:
\begin{align} \label{eq: phik}
\Psi_{k} = \Psi_{\lfloor k \rfloor + t}  =
& \mathcal{F}^{-1} \Bigg( \mathbb{I}_{\sqrt{k_x^2 + k_y^2 + k_z^2} < \lfloor k \rfloor} \cdot \mathcal{F}\{\phi\}(k_x, k_y, k_z) +  \left(1 - \mathbb{I}_{\sqrt{k_x^2 + k_y^2 + k_z^2} \ge \lfloor k \rfloor + 1}\right) \cdot \epsilon \notag \\
&\quad + \mathbb{I}_{\sqrt{k_x^2 + k_y^2 + k_z^2} \in [\lfloor k \rfloor, \lfloor k \rfloor + 1)} \cdot \left( \mu(t) \cdot \mathcal{F}\{\phi\}(k_x, k_y, k_z) + (1 - \mu(t)) \cdot \epsilon \right) \Bigg),
\end{align}
where $\mu(t)$ is a bump function such that $\mu(0) = \mu(1) = 1$ and $\mu'(0) = - \mu'(1)$. The antisymmetric property of $\mu'(t)$ ensures that $\Psi_{k}$ is differentiable from $k$ for all $\mathbb{R}^+$, allowing the flow matching loss and other gradient-based optimization techniques.
In~\Cref{eq: phik}, we have three components:
\begin{enumerate}[noitemsep,topsep=0pt]
    \item $ \mathbb{I}_{\sqrt{k_x^2 + k_y^2 + k_z^2} < \lfloor k \rfloor} $ applies to the amplitude components up to the integer part of $k$.
    \item $ \mathbb{I}_{\sqrt{k_x^2 + k_y^2 + k_z^2} \ge \lfloor k \rfloor + 1} $ applies noise padding to components beyond the next integer.
    \item $\mathbb{I}_{\sqrt{k_x^2 + k_y^2 + k_z^2} \in [\lfloor k \rfloor, \lfloor k \rfloor + 1)}$ performs linear interpolation of the intermediate amplitude components based on the current $t$.
\end{enumerate}

\textbf{Localized Vector Fields.}
Instead of directly modeling $\Psi_k$, we pivot our focus to its conditional gradient field, $\frac{d \Psi_k}{dk}$. By concentrating on the gradient field, we facilitate a dynamic view of how $\phi_k$ evolves with respect to $k$. To derive an analytical expression of $\frac{d\Psi_k}{dk}$ conditioned on a given instance pair of data and noise: $(\phi, \epsilon)$, in what follows, we assume that $\mathcal{F}$ is a linear transform. Then, following \Cref{eq: phik}, we have the conditional vector field:
\begin{equation} \label{eq:K_amplitude_vector_field}
\frac{d\Psi_k}{dk}(\phi, \epsilon) = \mathcal{F}^{-1} \Big(
    \mathbb{I}_{\sqrt{k_x^2 + k_y^2 + k_z^2}\in[[k],[k]+1)}\cdot \mu'(t) (\epsilon - \mathcal{F}\{\phi\}(k_x, k_y, k_z))
    \Big),
\end{equation}
for $k \in [\lfloor k \rfloor, \lfloor k \rfloor + 1)$ and $t= k - \lfloor k \rfloor$. Then, following \Cref{eq: conditional fm}, the training objective of \model{} is to learn the unconditional vector field in \Cref{eq: odeflow} by the conditional flow matching:
\begin{equation} \label{eq: loss}
\mathcal{L}_{\text{\model{}}} := \mathbf{E}_{\phi_0} \int_0^K \, d\phi_0 \, dk \, \left\| \frac{d \Psi_k}{dk} - v_k(\Psi_k, \theta) \right\|^2.
\end{equation}

By examining \Cref{eq: phik} closely, we observe that the vector field is naturally localized around a subset of points in the $K$-amplitude space that satisfy $\sqrt{k_x^2 + k_y^2 + k_z^2} \in [\lfloor k \rfloor, \lfloor k \rfloor + 1)$. This localization means that the reconstruction at any given $k$ primarily involves $K$-amplitude components within a narrow frequency band around $k$. Compared with the flow scheme in the pixel space, the $K$-amplitude in \model{} reduces the optimization complexity by restricting the conditional vector field to be within a sub-manifold for each $k$. This sub-manifold may potentially be of low dimensionality, allowing for more focused updates and reducing the optimization space's dimensionality at each step. We will investigate how this localized conditional vector field affects the generation path in \Cref{sec:discussion_complete}.

We can further generalize the interpolation interval from $(\lfloor k \rfloor, \lfloor k \rfloor+1)$ to $(k_m, k_n)$, where $k_m$ and $k_n$ are two integers such that $k_m < k_n$. This adjustment broadens the range for intermediate amplitude components from $\sqrt{k_x^2 + k_y^2 + k_z^2} \in [\lfloor k \rfloor, \lfloor k \rfloor + 1)$ to $\sqrt{k_x^2 + k_y^2 + k_z^2} \in [k_m, k_n)$. For example, for our experiments, we partition the $K$-amplitude into two or three parts.

\subsection{Examples of $K$-amplitude Transformation}
In this section, we illustrate how the $K$-amplitude decomposition $\mathcal{F}$ and its associated scaling parameter $k$ can be realized in various instances, extending beyond the (discrete) Fourier transform introduced in~\Cref{sec:background}. 

As we can see from \Cref{eq:signal_to_k_ampltitude}, all $K$-amplitude decompositions are achieved through expansion across a complete set of basis functions. However, the behavior of a $K$-amplitude decomposition (transform) can vary significantly depending on the choice of basis functions. Besides the Fourier transform introduced in~\Cref{sec:background}, we provide three examples of $K$-amplitude decomposition: Fourier transformation, Wavelet transformation, and PCA transformation.

\textbf{Fourier Transform.}
We have shown how to build the $K$-amplitude scaling parameter through the Fourier space in \Cref{sec:fourier_amplitude_decomposition_example}. In the discrete setting, the Fourier transform is realized by basis functions of the form $W_{N}^{kn} = e^{-j \frac{2\pi}{N} kn}$, where $N$ is the length of the sequential data. An effective $K$-amplitude decomposition exploits this structure by aligning with the inherent hierarchical structure of the data manifold. For example, if most of the energy or amplitudes are concentrated in the low-scaling range, the generative capability of the flow can be enhanced by allocating more steps or resources to these low frequencies. Conversely, fewer steps can be allocated to high frequencies that carry minimal mass or information. For the Fourier transform, this tendency is evident in the analysis of natural images, which often exhibit the celebrated $1/f$ spectrum phenomenon~\citep{weiss2007makes}. This phenomenon suggests that energy diminishes with increasing scaling parameter, meaning that low-scaling components hold the majority of the signal's information content. 

However, as we will see in \Cref{sec:results}, global Fourier transformations may not provide an optimal inductive bias, particularly when most data patterns are spatially local, {\eg}, facial expressions. To deal with this type of data, we consider wavelet bases that are not only scaling-localized but also spatially localized, allowing for a more nuanced representation of the data's structure.

\textbf{Wavelet Transform.}
Wavelet decomposition (transform) deals with data that are not only scaling-localized but also spatially localized. The scaling parameter of wavelet transform is closely related to the notion of multi-resolution analysis~\citep{mallat1989multiresolution}, which provides a systematic way to decompose a signal into approximations and details at successively finer scales. This hierarchical decomposition is achieved through a set of scaling functions $\omega(x)$, and wavelet functions $\psi(x)$, which together serve as basis functions for the wavelet transformation. More precisely, in the wavelet transform, a signal $f(t)$ is expressed as a sum of scaled and translated versions of these basis functions times the corresponding coefficients $c$ and $d$:
\begin{equation} \label{eq: wavelet-decomposition}
f(t) = \sum_{j} c_{k_0, j} \omega_{k_0, j}(t) + \sum_{k \geq k_0} \sum_{j} d_{k, j} \psi_{k, j}(t),
\end{equation}
where $\omega_{k_0, j}(t)$ and $\psi_{k, j}(t)$ are the scaled and translated scaling and wavelet functions, respectively. The index $j$, which originally denotes the translation parameter, groups the basis within each fixed scaling parameter $k$ naturally. Let $\phi_k : = \sum_{j} d_{k, j} \psi_{k, j}$ for $k>k_0$ and $\phi_k := \sum_{j} c_{k_0, j} \omega_{k_0, j}$ for $k= k_0$, then eq. \ref{eq: wavelet-decomposition} is just one realization of $K$-amplitude decomposition.

In this article, we employ the discrete version of wavelet transform (DWT) as our $K$-amplitude transformation $\mathcal{F}$, which shares the linearity property with the Fourier transform with a bounded scaling parameter $k$, providing a structured yet flexible means of decomposing discrete data. 

\textbf{Date-dependent PCA Transform.}
Note that Fourier and wavelet decompositions are nonparametric k-amplitude decompositions that are independent of data. While these transformation methods are powerful, we also aim to find data-dependent decompositions that can capture common characteristic features specific to a given dataset. This motivation leads to principal component analysis (PCA), a technique widely used for the low-dimensional approximation of the data manifold \citep{izenman2012introduction}.\looseness=-1

\begin{figure}[t]
\centering
    \includegraphics[width=.8\linewidth]{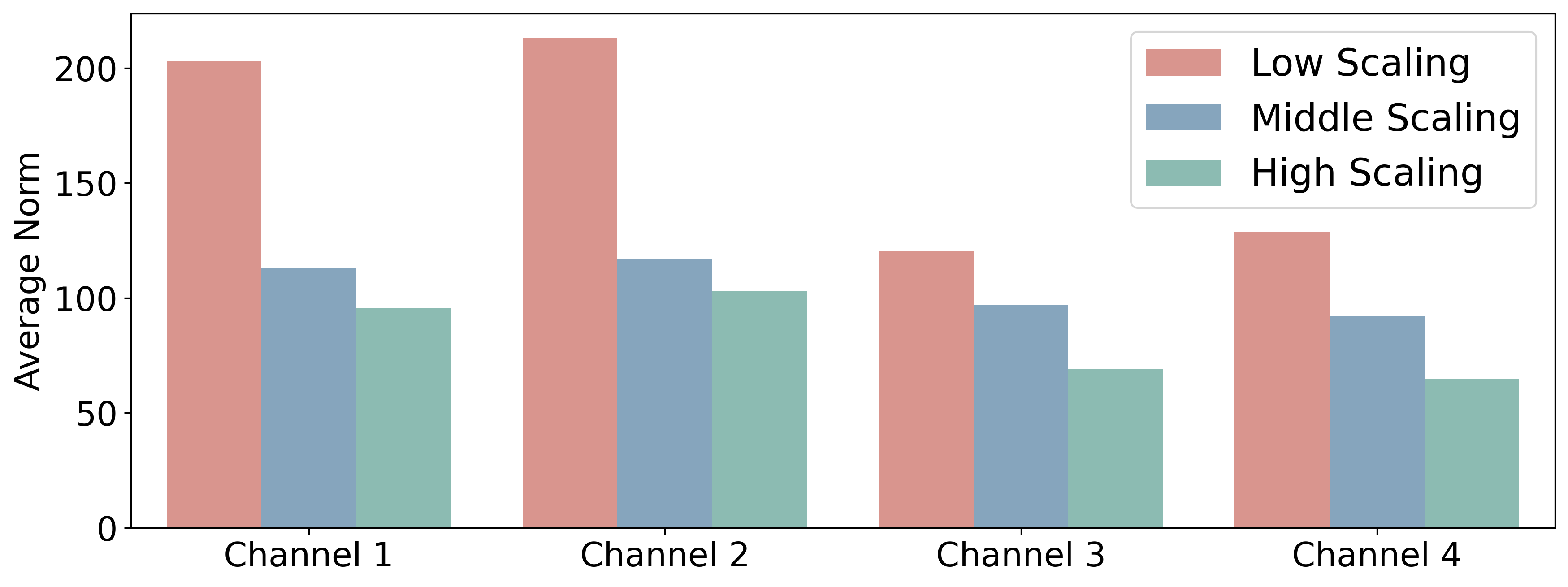}
    \vspace{-2ex}
    \caption{On the low-scaling hypothesis. \small The graph illustrates the relative norm distribution for each scaling component as defined by the wavelet decomposition in the latent space. It can be observed that the low-scaling component exhibits a significantly higher norm (energy), nearly twice that of the high-scaling component.}
    \label{fig:low_frequency}
\end{figure}

\textbf{$K$-amplitude Decomposition As A good Inductive Bias.}
From a data modeling perspective, it is valuable to study the statistics of data distribution across scalings, as defined by the specific $K$-amplitude decomposition we utilize. If the data distribution does not exhibit \model{} scaling inhomogeneity, then all scalings should be treated equally, providing no justification for using a scaling-split generation path. As to latent data modeling, such as the latent space of an autoencoder, which is our main focus, we statistically analyze the mean norm of each scaling band across images in \Cref{fig:low_frequency}. Obviously, we find that even in the compressed latent space, the mean norm of each scaling band decreases from low to high scalings. From the perspective of approximation error and model complexity, it is advantageous to allocate more refined sampling steps (or more model parameters) to lower scalings, as they contain more energy. On the other hand, pathological medical imaging data \citep{chu2024improving} may place more emphasis on the reconstruction of high-frequency components. In such cases, we need to allocate more computational resources to the high-scaling part of the \model{}. Overall, our method allows the model to better capture significant features and maintain fidelity in the generated outputs.

\begin{algorithm}[t]
\caption{Training of \model{}.}
\label{alg:training}
\begin{algorithmic}
\REQUIRE Scaling parameter $k$ with maximum $k_{max}$, \model{} transform $\mathcal{F}$, inverse transform $\mathcal{F}^{-1}$,  noise distribution $p$, target distribution $q$
\STATE Normalize $k$ to be in $[0,1]$:  $k \leftarrow k/k_{max}$
\STATE Initialize parameters $\theta$ of $v_k$
\WHILE{not converged}
    \STATE Sample scaling parameter $k \sim \mathcal{U}(0, 1)$
    \STATE Sample training example $\phi \sim q$, sample noise $\epsilon \sim p$
    \STATE Calculate current flow position $\Psi_k$ according to \model{} transform $\mathcal{F}$, $\mathcal{F}^{-1}$ and \Cref{eq: phik}
    \STATE Calculate the conditional vector field $\dot{\Psi}_k$ according to  $\mathcal{F}$, $\mathcal{F}^{-1}$ and \Cref{eq:K_amplitude_vector_field}
    \STATE Calculate the objective $\ell(\theta) = \|v_k(\Psi_k;\theta) - \dot{\Psi}_k\|_g^2$, following \Cref{eq: loss}
    \STATE $\theta = \text{optimizer\_step}(\ell(\theta))$
\ENDWHILE
\end{algorithmic}
\end{algorithm}

\subsection{Practical Implementation and Discussion}
The overall structure of \model{} is agnostic to the neural network architecture (for training the vector field), meaning that classical model architectures, such as U-net~\citep{song2020score} and vision transformers~\citep{peebles2023scalable}, which are commonly used for training ordinary continuous normalizing flows or diffusion models, can be directly applied to \model{}. This adaptability ensures that existing computational investments in these architectures can be effectively leveraged, providing a seamless transition to incorporating $K$-amplitude-based methods.

\textbf{A Flexible Plug-In Version.}
To integrate our method into these existing models, we only introduce one targeted modification: replacing the time-embedding module with a $K$-amplitude-embedding module. Specifically, the time input of the time-embedding module in the diffusion transformer (or U-net) is substituted by the scaling parameter $k$. This substitution enables the \model{} to leverage scaling information directly (especially the bump function), aligning with the principles of $K$-amplitude decomposition while preserving the original architecture's overall structure.

\textbf{Remarks.}
Despite this model-agnostic nature, the unique $K$-amplitude localization property of \Cref{eq:K_amplitude_vector_field} offers an opportunity to design more efficient models. For instance, consider points that lie outside the support of function $\mathbb{I}_{\sqrt{k_x^2 + k_y^2 + k_z^2} \in [\lfloor k \rfloor, \lfloor k \rfloor + 1)}$. In these regions, their derivative remains zero, indicating that they do not contribute to the optimization process for the corresponding scaling band. This selective activation allows us to focus computational efforts solely on the values within the support of the indicator function, $\mathbb{I}_{\sqrt{k_x^2 + k_y^2 + k_z^2} \in [\lfloor k \rfloor, \lfloor k \rfloor + 1)}$. By doing so, the values outside this region can be treated as static conditions, providing a fixed context.

\section{Discussion} \label{sec:discussion_complete}

\subsection{From Conditional to Unconditional Path in \model{}}
\label{sec: From Conditional to Unconditional Path}
In~\Cref{sec:method}, our frequency-localized path is defined at the conditional level ($\frac{d\Psi_k}{dk}( \phi,\epsilon)$ ) , and it is only related to the unconditional vector field ($v_k(\Psi_k, \theta)$ in \cref{eq: loss}) through the equivalence of conditional flow matching and unconditional flow matching at the loss level~\citep{lipman2022flow}. In this section, we try to study the splitting property of the unconditional $K$-amplitude vector field. 

By the $K$-amplitude decomposition, the transformed data probability $p_{data}$ satisfies the telescoping property:
\begin{equation}
p_{data}  = p(k_0)p(k_1 | k_0) \dots p(k_{max} |k_{max} -1,  \dots, k_0),    
\end{equation}
with $k_0$ and $k_{max}$ denoting the lowest and highest scaling. Then, according to the definition of our proposed \model{} $\Psi_k$, the interpolated probability at scaling step $t$ is also localized:
\begin{equation} \label{eq:splitting}
p_t(\cdot)  = p(k_0)\cdots p_t( \cdot | \lfloor k \rfloor , \dots, k_0) p_{\epsilon}(\lfloor k \rfloor +1) \cdots p_{\epsilon}(k_{max} |k_{max} -1,  \dots, k_0) ,
\end{equation}
where $p_{\epsilon}$ denotes the distribution of the initial noise and $t \in [\lfloor k \rfloor, \lfloor k \rfloor+1)$. Combining \Cref{eq:splitting}, the localization property of the bump function, and Lemma 1 of \citep{zheng2023guided}, the unconditional  vector field has an explicit form:
$v_t(\Psi_k) = a_t \cdot \Psi_k + b_t \nabla \log p_t(\Psi_k),$
where $a_t$ and $b_t$ are hyper-parameters determined by the bump function we choose. 

\textbf{Noise Splitting.}
A key characteristic of flow models is their deterministic nature after the initial noise sampling. Specifically, once the initial noise is sampled, the flow follows a fixed path to generate the final data sample. According to \Cref{eq:splitting}, during scaling step t: (1) the scaling components below $\lfloor k \rfloor$ remain unchanged; (2) the scaling components above $\lfloor k \rfloor$ remain unchanged; (3) The distribution of higher scaling components maintains the same characteristics as their initial noise distribution.

By these observations, we now investigate how segmented initial noise in the \model{} space influences the final output of the \model{} flow. Suppose we discretize scaling parameter $k$ into two parts: $\mathcal{F}\{\Psi_k\} = \{\phi_{\text{low}}(k), \phi_{\text{high}}(k)\}$. When flowing along the low-scaling component, the vector field $v_k$ can be re-expressed in a conditional form:
\begin{equation}
v_k(\Psi_k) = v_k(\phi_{\text{low}}(k), c)
\end{equation}
where constant $c$ represents the (static) initial noise for the high-scaling part. This noise-conditioned property in the k-amplitude domain leads us to explore whether fixing the high-scaling noise and altering the low-scaling noise allows for unsupervised editing of relative low-scaling semantics in an image. Indeed, we observed this phenomenon, the qualitative results will be discussed in section \ref{sec: editing}.

From \Cref{fig:06_steerable_generation_results}, we observe that a targeted common high-scaling initial noise guides our \model{} flow toward generating human faces with similar detail but varying low-level content. See the experiment section for a more detailed analysis.

\begin{figure}[t]
    \centering
    \includegraphics[width=.8\linewidth]{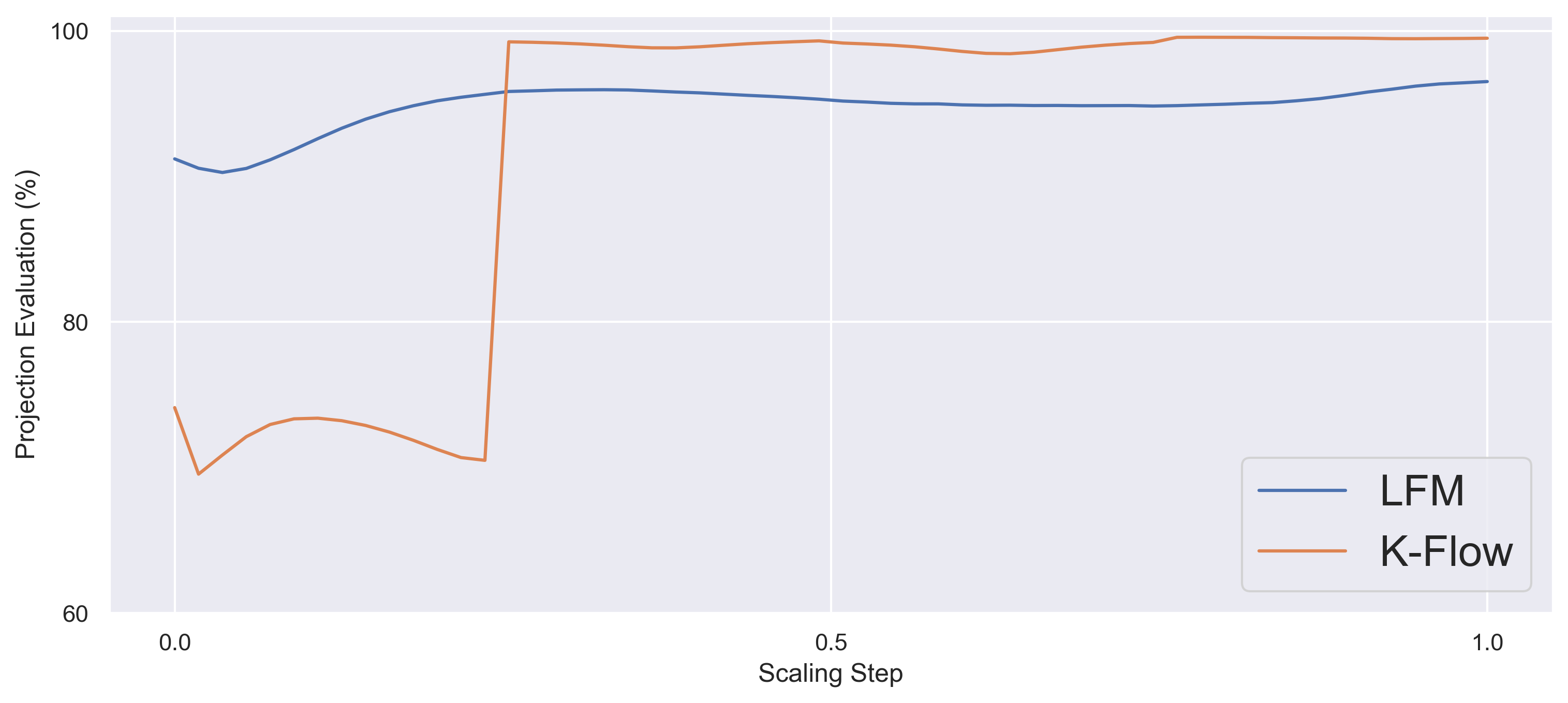}
    \vspace{-2ex}
    \caption{Projection Error Comparison with Different Models. 
    \small
     The graph illustrates the PCA projection errors of two models throughout the entire flow process, with distinct segments marked by dashed lines. The red and blue lines represent the original latent flow matching (LFM) and the \model{} with two amplitude components, respectively. The projection error is quantified by the reconstruction error for each generation step from the PCA compression, using the first two principal components. Owing to the scaling-aware nature of our flow, the low-amplitude portion (the initial part of the curve) resides in a relatively high-dimensional space, resulting in higher projection errors for the two-dimensional PCA projection.
    }
    \label{fig:projection_error}
\end{figure}

\subsection{The Effect of Scaling Step $k$ for Image Reconstruction}
\model{}'s ability to leverage the low-dimensional structure of data is primarily enabled by its \model{} localization property. This enables a strategic path through low-dimensional spaces, which can be directly compared with the generation path of conventional flow models. In our model, this path incorporates an explicit frequency hierarchy, which hypothesizes that the low-frequency components - concentrated in the earlier stages of the model - may share more dimensions in common, particularly from a semantic perspective, than the high-frequency components positioned later in the generative process. Conversely, an ordinary flow model may exhibit a more uniform distribution of dimensionality across the entire generative path.

Motivated by this hypothesis, we conduct a case study using PCA to approximate the dimension of the generation trajectory $\{ \Psi_k \}_{k=k_0}^{k_{max}}$. As illustrated in \Cref{fig:projection_error}, we measure how closely the dimension of the generation path aligns with a two-dimensional subspace spanned by the first two components of the model's PCA decomposition, denoted by $\{ \Tilde{\Psi}_k \}_{k=k_0}^{k_{max}}$. Inspired by \citep{zhou2024fast}, the reconstruction ratio is defined by 
$1 - \|\Psi_k - \Tilde{\Psi}_k\|_2/ \|\Psi_k \|_2$.
In other words, a higher value of the reconstruction ratio indicates that the model's dimension is closer to two.
Therefore, the trend of the error curve with respect to the scaling parameter $k$ reveals a distinct separation in the effective dimension between low- and high-scaling components. Evidently, the low-scaling segments display more semantic consistency and thus, occupy a larger dimensional space, whereas the high-scaling segments converge to a more confined or lower-dimensional structure.

It is important to note that this exploration into the dimensionality of generative paths is practically meaningful. Previous study \citep{zhou2024fast} has shown that the effectiveness of distilling a generative model with fewer steps from a pre-trained diffusion model theoretically depends on the model's dimensionality at each step, as informed by the high-dimensional Mean Value Theorem.
The observations from \Cref{fig:projection_error} provide empirical support for this concept. Specifically, the ability of \model{} to maintain a lower-dimensional structure in high-scaling components suggests a promising approach for fast sampling distillation methods.

\begin{figure}[tb]
    \centering
    \includegraphics[width=.8\linewidth]{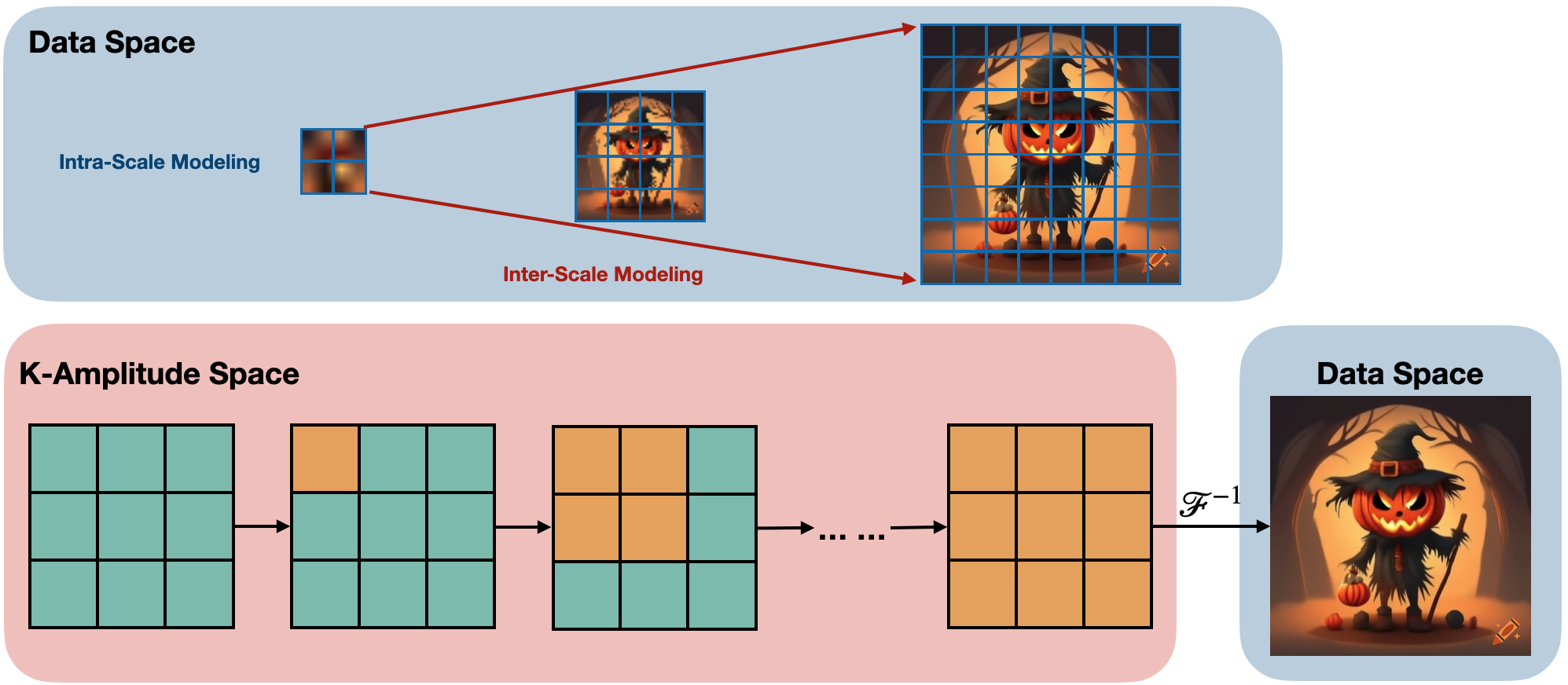}
    \vspace{-2ex}
    \caption{\small Comparison of multi-scale modeling: pixel data space and K-Amplitude space.}
    \label{fig:related_work_comparison}
\end{figure}

\subsection{Related Work Discussion}
The field of generative modeling has seen significant advancements in recent years, driven by a variety of frameworks, including adversarial generative networks (GAN) \citep{goodfellow2014generative}, variable autoencoders (VAE) \citep{kingma2013auto}, and normalizing flows \citep{papamakarios2021normalizing}. In this work, we focus on continuous normalizing flow generative models \citep{chen2018neural}, with particular emphasis on the conditional flow matching training scheme, which originates from the denoising score matching training framework \citep{vincent2011connection}.

Both diffusion models and continuous flow matching models aim to lower the complexity of directly optimizing the log-likelihood of data by introducing an additional stochastic path. However, as proved in \citep{lavenant2022flow}, the canonical path for diffusion models and rectified flows is not optimal. This realization motivates our introduction of frequency decomposition as a key design element in generative models.

By breaking down the formula of our \model{} vector field with respect to the scaling parameter $k$, we can summarize three successful factors as general principles for (frequency) scaling modeling.
\begin{itemize}[noitemsep,topsep=0pt]
    \item A good $K$-amplitude decomposition can leverage the problem's inherent biases towards certain scaling bands. For generative tasks, it is crucial that all \model{} bands are effectively modeled to ensure the generation of high-quality, controllable outputs. In addition, the computational resources required may vary between different scales, thus necessitating careful consideration of resource allocation.
    \item  Modeling within each scaling component, which is formulated in our \model{}-localized vector fields.
    \item Modeling bridges along different scalings, which is achieved through our flow ODE and the (time) \model{} embedding block for the U-Net or DIT architecture.
\end{itemize}
This approach to inter- and intra-modeling for $K$-amplitude is also applicable to scenarios emphasizing certain frequencies or scalings. For instance, \citep{li2024generative} enhanced oscillatory motion control in video generation by discarding the high-frequency component of the Fourier decomposition. As discussed in \Cref{sec:method}, the scaling parameter of spatially localized wavelet (multi-resolution) decomposition is closely linked to image resolution. Notable contributions in this domain include \citep{atzmon2024edify} and \citep{lei2023pyramidflow}, which introduced a multi-stage resolution for fine-grained editing, and \citep{jin2024pyramidal}, which concentrated on efficient video generation.

In related research on auto-regressive modeling, \citep{mattar2024wavelets} presented wavelets as an effective auto-regressive unit, while \citep{tian2024visual} focused on the scale as a key element for image auto-regression. A significant example is \citep{phung2023wavelet}, which transitioned the latent space from pixel to wavelet space for generative models using wavelet diffusion. However, their method employed the same conditional noising schedule for score matching as traditional diffusion models. In contrast to their approach, our proposed \model{} integrates wavelet decomposition as a multi-channel module within the neural network architecture for training diffusion models. Additionally, our work extends the notion of wavelet space to the more general $K$-amplitude space. 

We also want to highlight another research line that has recently caught the attention is the auto-regressive over the pixel space for image generation. One classic work is VAR~\cite{tian2024visual}. It introduces a hierarchical density estimation paradigm that models images in a coarse-to-fine manner across multiple resolutions and models the data distribution in an auto-regressive manner. In contrast, our proposed \model{} integrates the flow paradigm for density estimation and leverages the $K$-amplitude space as a stronger inductive bias, as illustrated in \Cref{fig:related_work_comparison}.

\textbf{Summary.}
In summary, \model{} is a more general framework, with its three key factors potentially benefiting generation-related tasks like super-resolution and multi-resolution editing. For example, \citep{liucosae} utilized a learnable Fourier transform to construct a harmonic module in the bottleneck layer of an autoencoder. We provide a comprehensive list of related works in \Cref{sec: related work}.

\begin{table}[t]
\centering
\caption{Comparison among PCA, contrastive, and generative SSL.}
\vspace{-2ex}
\begin{adjustbox}{max width=\textwidth}
\begin{tabular}{l rrr}
\toprule
    & Basis Learning & Reconstruction Learning \\
\midrule
PCA SSL & Non-parameterized, Determined By Data & Parameterized\\
Contrastive SSL & Parameterized & N/A\\
Generative SSL & Parameterized & Parameterized\\
\bottomrule
\end{tabular}
\end{adjustbox}
\end{table}

\subsection{Connecting \model{} with SSL Representation and Generation}
From the above discussion, we have seen how pretrained vision models leverage the sparsity and locality of natural data in various $K$-amplitude domains for perception and generation-based tasks. In the realm of unsupervised learning, \citep{liu2022molecular,liu2024symmetry,chen2024deconstructing} explore whether generative-based representations, particularly those derived from denoising diffusion models, can achieve parity with contrastive-based representation learning methods for downstream tasks. A key observation from their findings~\cite{chen2024deconstructing}, which aligns with our approach of employing $K$-amplitude decomposition (the PCA instance), is the revelation that the most powerful representations are obtained through denoising within a latent space, such as the compressed PCA space. Another merit of PCA is that denoising along the PCA directions can achieve faster convergence for denoising, which is revealed in \citep{du2023flexible}.

To transition from unsupervised representation learning to real data generation, incorporating all $K$-amplitude scalings is essential. Rather than compressing or amplifying specific scaling bandwidths, generative tasks require novel organization or ordering of all frequencies. Besides our flow-based frequency generation approach, \citep{tian2024visual} connects different scales (which can be interpreted as the wavelet $K$-amplitudes) using residual connections with an auto-regressive training objective. Residual connections, as a discretization of ordinary differential equations (ODEs) proposed in \citep{weinan}, suggest that \citep{tian2024visual}'s approach can be seen as a special discretization of our \model{} with a flexible flow matching training objective.

\section{Experiments} \label{sec:results}
Our proposed \model{} is designed to: (1)  Ensure stable training by leveraging the inherent stability of conditional vector field matching in (latent) flow matching frameworks. (2) Incorporate amplitude-adapted generation paths, enabling more controlled and interpretable generative processes through $K$-amplitude decomposition and flow matching. To empirically demonstrate these attributes, we evaluate our \model{} in multiple tasks, including unconditional image generation, class-conditioned image generation, and three ablation studies.

\subsection{Image Unconditional Generation}
The first task is to generate random samples after fitting a target data distribution, which is typically concentrated around a low-dimensional sub-manifold within the ambient space.

\textbf{Dataset and Metrics.}
We conduct experiments on the CelebA-HQ \citep{karras2017progressive} dataset with the resolution of \(256 \times 256\) and LSUN Church \cite{yu2015lsun} dataset with the resolution of $256 \times 256$. To evaluate the performance of our proposed method, we employ two metrics: the Fréchet Inception Distance (FID) \citep{heusel2017gans}, which evaluates the quality by measuring the statistical similarity between generated and real images, and Recall \citep{kynkaanniemi2019improved}, which measures the diversity of the generated images.

\textbf{Results.}
\Cref{tab:unconditional_generation_celeba} summarizes the comparison between our proposed \model{} model and other generative models. For a fair comparison, both the baseline ordinary flow matching \citep{dao2023flow} and our \model{} flow utilize the same VAE's latent from \citep{rombach2022high} and the Diffusion Transformer with the same size ({\eg}, DIT L/2 \citep{peebles2023scalable}) as the backbone model. We can observe that (1) \model{} achieves the best performance in FID, especially w/ the db6-based wavelet \model{}. (2) Although the latent diffusion model \citep{rombach2022high} gets the highest score in Recall (diversity), the Fourier and PCA-based \model{} is comparable with the ordinary latent flow matching. 

\Cref{tab:unconditional_generation_celeba} summarizes the results on LSUN Church. We test our \model{} with two and three scaling components using the db6 wavelet $K$-amplitude transform, and we find that the three scaling components version achieves the best quantitative results in terms of FID and Recall.

\begin{table}[t]
    \begin{minipage}[t]{0.49\textwidth}
    \centering
    \caption{\small Unconditional generation on CelebA-HQ 256.} \label{tab:unconditional_generation_celeba}
    \vspace{-2ex}
    \begin{adjustbox}{max width=\linewidth}
    \begin{tabular}{lcc}
        \toprule
        \textbf{Model} & \textbf{FID$\downarrow$} & \textbf{Recall$\uparrow$} \\
        \midrule
        \model{}, Fourier-DiT L/2 (Ours) & 5.11 & 0.47 \\
        \model{}, Wave-DiT L/2 (Ours) & \textbf{4.99} & 0.46 \\
        \model{}, PCA-DiT L/2 (Ours) & 5.19 & 0.48 \\
        LFM, ADM \citep{dao2023flow} & 5.82 & 0.42 \\
        LFM, DiT L/2 \citep{dao2023flow} & 5.28 & 0.48 \\
        FM \citep{lipman2022flow} & 7.34 & - \\
        \cmidrule(r){1-3}
        LDM \citep{rombach2022high} & 5.11 & \textbf{0.49} \\
        LSGM \citep{vahdat2021score} & 7.22 & - \\
        WaveDiff \citep{phung2023wavelet} & 5.94 & 0.37 \\
        DDGAN \citep{xiao2021tackling} & 7.64 & 0.36 \\
        Score SDE \citep{song2020score} & 7.23 & - \\
        \bottomrule
    \end{tabular}
    \end{adjustbox}
    \end{minipage}%
\hfill
    \begin{minipage}[t]{0.49\textwidth}
    \centering
    \caption{\small Unconditional generation on LSUN Church 256.} \label{tab:lsun-church-results}
    \vspace{-2ex}
    \begin{adjustbox}{max width=\linewidth}
    \begin{tabular}{lcc}
    \toprule
    \textbf{Model} & \textbf{FID $\downarrow$} & \textbf{Recall $\uparrow$} \\
    \midrule
        \model{}, two scales (Ours) & 5.37 & 0.47 \\
        \model{}, three scales (Ours) & \textbf{5.19} & \textbf{0.49} \\
        \midrule
        LFM (ADM) & 7.7 & 0.39 \\
        LFM (DiT L/2) & 5.54 & 0.48 \\
        FM  & 10.54 & - \\ \hline
        LDM  & 4.02 & 0.52 \\
        WaveDiff  & 5.06 & 0.40 \\
        DDPM  & 7.89 & - \\
        ImageBART  & 7.32 & - \\
    \bottomrule
    \end{tabular}
    \end{adjustbox}
    \end{minipage}
\end{table}

\subsection{Image Class-conditional Generation}
In this section, we explore how $K$-amplitude decomposition behaves when the generation path is conditioned on class labels, where the class label ({\eg}, dog, cat, fish, etc) delegates the low-scaling information of each image. This investigation could potentially pave the way for multi-scaling control, where different scaling components are influenced by specific caption information.

\textbf{Dataset and Metric.} We use ImageNet as the middle-size conditional generation dataset~\citep{deng2009imagenet}. Beyond evaluating the unconditional FID for the ImageNet dataset, we are interested in studying how the class control interacts with scaling generation in a quantitative manner.

We propose using the class-aware FID metric, defined as follows:
\begin{equation}
\text{FID}_{\text{class-conditional}} = \mathbb{E}_{c \sim p(c)} \left[ \text{FID}(c) \right]
\end{equation}
where for each class $c$, the FID is calculated by:
\begin{equation}
\text{FID}(c) := \text{FID}(X_r^c, X_g^c) = \lVert \mu_r^c - \mu_g^c \rVert^2 + \text{Tr}(\Sigma_r^c + \Sigma_g^c - 2(\Sigma_r^c \Sigma_g^c)^{1/2}).
\end{equation}
Here, $X_r^c$ and $X_g^c$ denote the real and generated data subsets for class $c$, respectively.  Based on $\text{FID}(c)$, the Class-Dropping-Ratio (CDR) is defined by
\[
\text{CDR} := \mathbb{E}_{c \sim p(c)} \left[\frac{\text{FID}_{\text{bef}}(c)}{\text{FID}_{\text{aft}}(c)}\right],
\]
where $\text{FID}_{\text{bef}}$ denotes the FID calculated for the flow model carried with the class condition for the whole process, and $\text{FID}_{\text{aft}}$ denotes the FID calculated for the flow model carried with the class condition for only a subprocess (we keep the initial 30\% of the inference time for the experiment). In practice, instead of computing the expectation over the entire class distribution $p(c)$, we randomly select 5 classes out of the total 1000 classes for evaluation.

\begin{wraptable}[11]{R}{0.48\textwidth}
\vspace{-3ex}
\caption{\small Class-conditional generation on ImageNet.} \label{tab:conditional_generation_imagenet}
\vspace{-2ex}
\begin{adjustbox}{max width=\linewidth}
\begin{tabular}{lccc}
    \toprule
    \textbf{Model} & \textbf{FID$\downarrow$} &
    \textbf{CDR$\downarrow$} &
    \textbf{Recall$\uparrow$} \\
    \midrule
    \model{}, Wave-DiT L/2 (Ours) & 17.8 & - & 0.56 \\
     + cfg=1.5 & 4.49 & - & 0.44 \\
     \model{}, Fourier-DiT L/2 (Ours) & 13.5 &-& 0.57 \\
    + cfg=1.5 & \textbf{2.77} & \textbf{1.49} & 0.45 \\
     LFM, DiT L/2 & 14.0 & - & 0.56 \\
    + cfg=1.5 & 2.78 & 3.25 & 0.42 \\
    \cmidrule(r){1-4}
    LDM-8 \citep{rombach2022high} & 15.51 & - & \textbf{0.63} \\
    LDM-8-G & 7.76 & - & 0.35 \\
    DiT-B/2 \cite{peebles2023scalable} & 43.47 &- & - \\
    \bottomrule
\end{tabular}
\end{adjustbox}
\end{wraptable}

\textbf{Results.}
The results are presented in \Cref{tab:conditional_generation_imagenet}. Our primary focus for the FID metric is the classifier-free guidance inference method applied to flow matching models. The data indicates that \model{} achieves results comparable to LFM. In terms of the recall metric, which assesses the diversity of the generated distribution, our model outperforms the standard LFM. This improvement may be attributed to the fact that the inference path of \model{} includes a greater number of dimensions during the low-scaling period, as discussed in \Cref{sec: From Conditional to Unconditional Path}.
Given that the inference path of \model{} accommodates the $K$-amplitude scalings, we anticipate that omitting the class label (a low-scaling caption) in the high-scaling segments will not substantially impact FID. Our observations confirm this expectation: the conditional discrimination ratio of our model is close to one, indicating a balanced performance. In contrast, the CDR of the conventional LFM is significantly higher, suggesting a discrepancy in performance under these conditions. For the qualitative analysis, see \Cref{sec:class-conditional_generation}. This preliminary exploration suggests that our proposed \model{} has the potential to allocate computational resources more efficiently by leveraging the correlation between scaling parameter $k$ and captions.\looseness=-1

\begin{figure}[t]
    \centering
    \includegraphics[width=.8\linewidth]{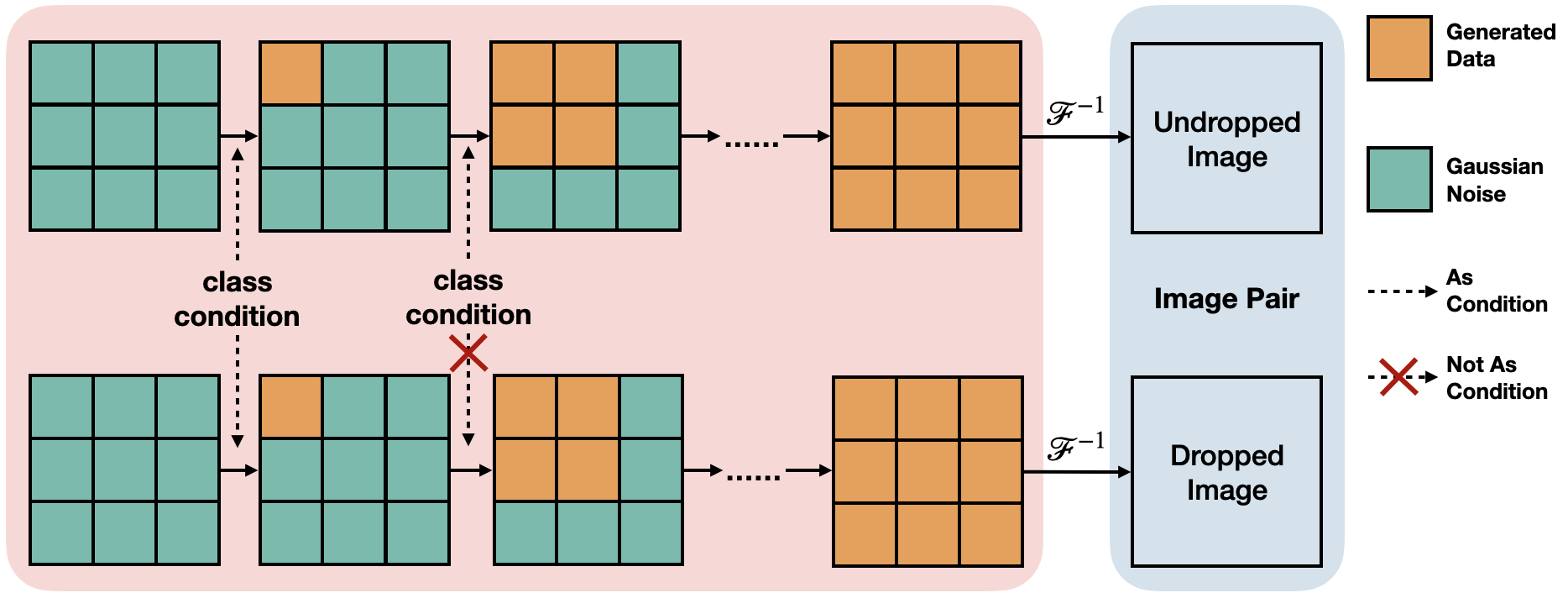}
    \vspace{-2ex}
    \caption{\small Pipeline of controllable class-conditional generation. }
    \label{fig:05_class_conditional_generation_pipeline}
\end{figure}
\begin{figure}[t]
    \centering
    \includegraphics[width=.8\linewidth]{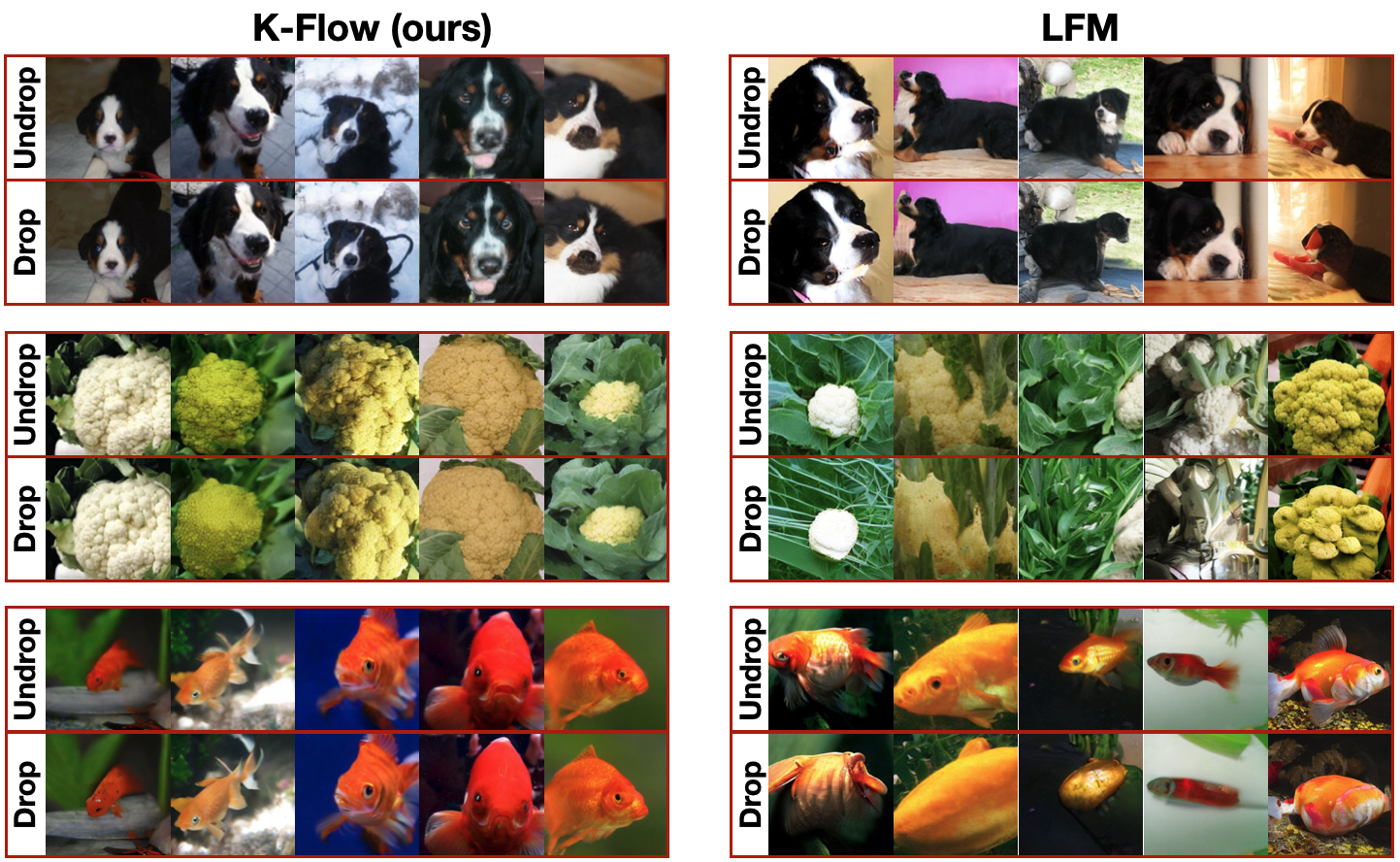}
    \vspace{-2ex}
    \caption{\small Results of controllable class-conditional generation. `Drop' means we drop the class conditions during the last 70\% scaling steps, while `undrop' means we keep the condition all the time.}
    \label{fig:05_class_conditional_generation_results}
\end{figure}

\subsection{Image Controllable Class-conditional Generation} \label{sec:class-conditional_generation}
The latent flow matching model can implicitly learn low- and high-resolution features~\citep{dao2023flow}, but the boundary between each resolution is vague, and we cannot explicitly determine which timestep in the inference process corresponds to a specific resolution or frequency. In comparison, our proposed wavelet-based \model{} enables finer-grained controllable generation. As shown in \Cref{fig:05_class_conditional_generation_pipeline,fig:05_class_conditional_generation_results}, when we drop the class conditions during the last 70\% scaling steps of the inference process, \model{} can effectively preserve high-frequency details, whereas the ordinary latent flow tends to blur the entire image.

\subsection{Image Scaling-controllable Generation} \label{sec: editing}
As discussed in \Cref{sec:discussion_complete}, our unconditional k-amplitude generation path reveals the $k$ scaling separation phenomenon. This allows us to control initial noise at each scaling level, enabling unsupervised editing of different scaling components.

\begin{figure}[t]
    \centering
    \includegraphics[width=.8\linewidth]{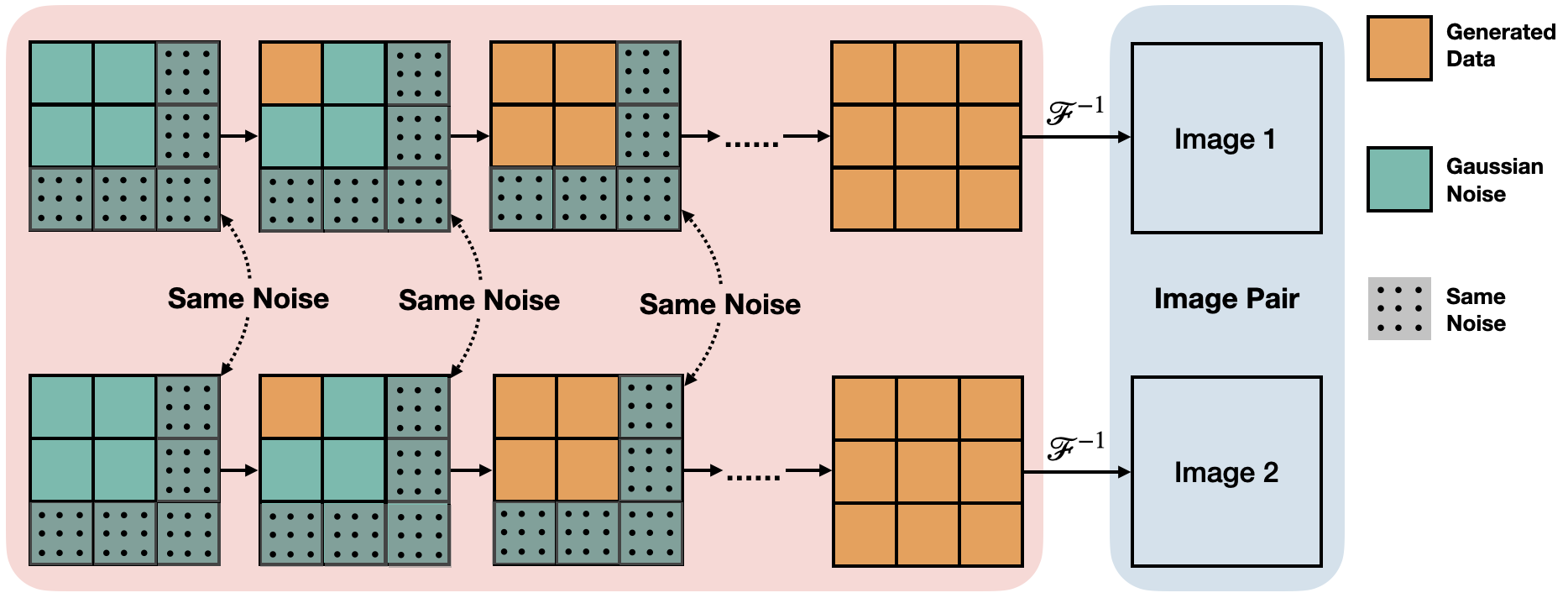}
    \vspace{-2ex}
    \caption{\small Pipeline of scaling-controllable generation (low scaling).}
    \label{fig:06_steerable_generation_pipeline}
\end{figure}

\textbf{Preserving High Scaling, Modifying Low Scaling.}
This scaling-controllable generation pipeline is illustrated in \Cref{fig:06_steerable_generation_pipeline}. It involves sampling multiple images while ensuring that the noise in the high-scaling components remains consistent across all samples. In scaling-controllable image generation, the goal is to maintain consistency in the high-scaling details while allowing variations in the low-scaling context among the generated images, thus this allows \model{} to achieve unsupervised steerability in a finetuning-free manner.

\begin{figure}[t]
    \centering
    \includegraphics[width=.8\linewidth]{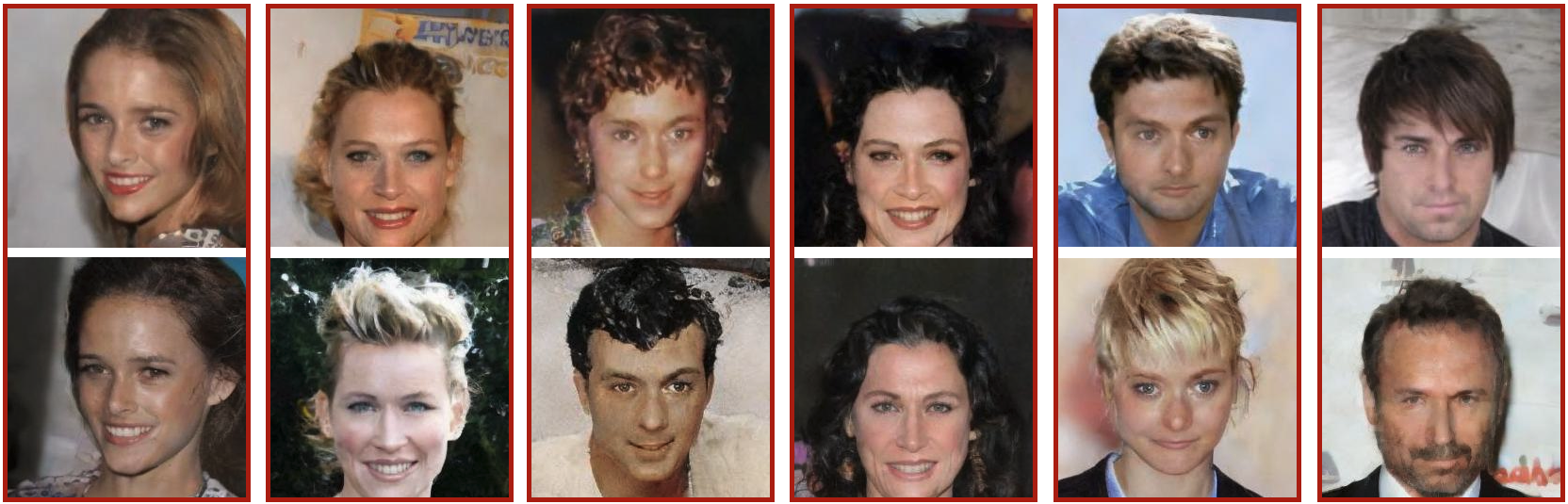}
    \vspace{-2ex}
    \caption{\small Results of scaling-controllable generation. We display six pairs of images, where each pair of images preserves the high scaling and differs in the low scaling.
    }
    \label{fig:06_steerable_generation_results}
\end{figure}

The results on CelebA are presented in \Cref{fig:06_steerable_generation_results}, where we apply a pretrained Daubechies wavelet (db6-based) \model{}. It can be observed that facial details, such as eyes, smiles, noses, and eyebrows, remain consistent within each group of images. In contrast, the low-scaling components, including background, gender, age, and hairstyle, vary across the images within the same group.

\begin{figure}[t]
    \centering
    \includegraphics[width=.8\linewidth]{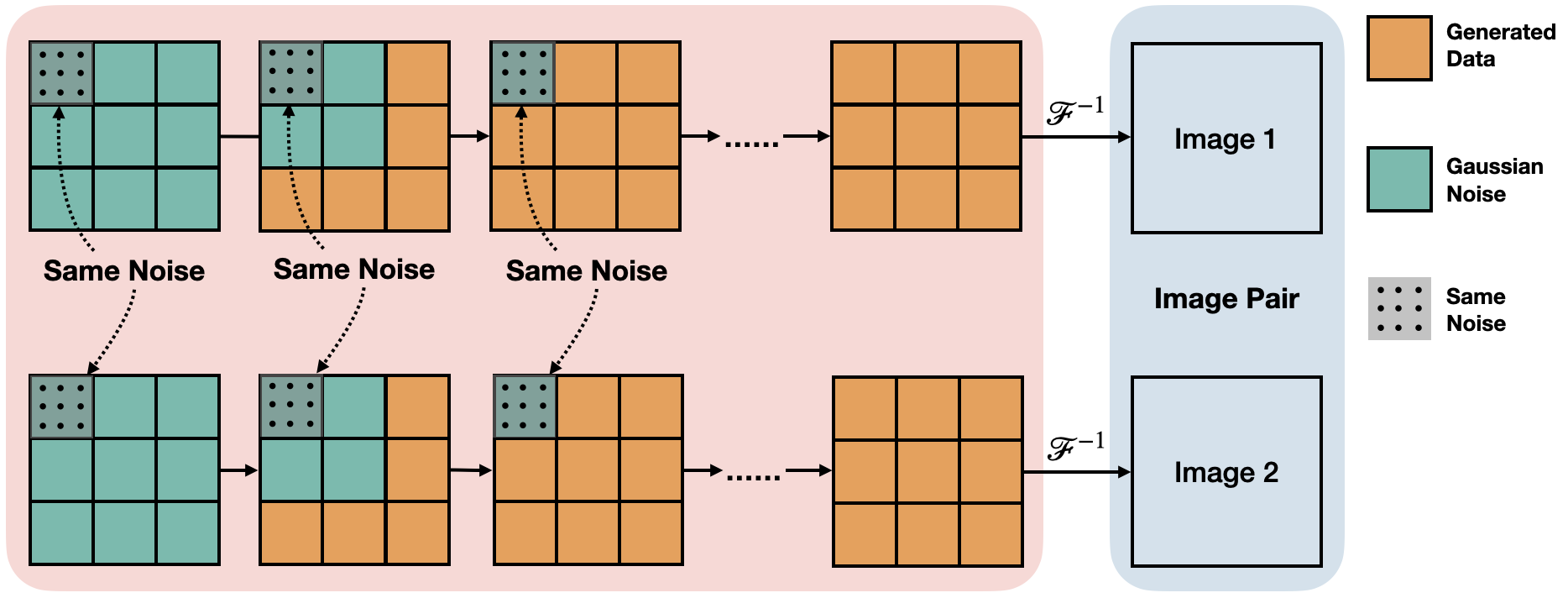}
    \vspace{-2ex}
    \caption{\small Pipeline of scaling-controllable generation (high scaling).}
    \label{fig:07_steerable_generation_pipeline}
    \vspace{-2ex}
\end{figure}

\begin{figure}[t]
    \centering
    \includegraphics[width=.8\linewidth]{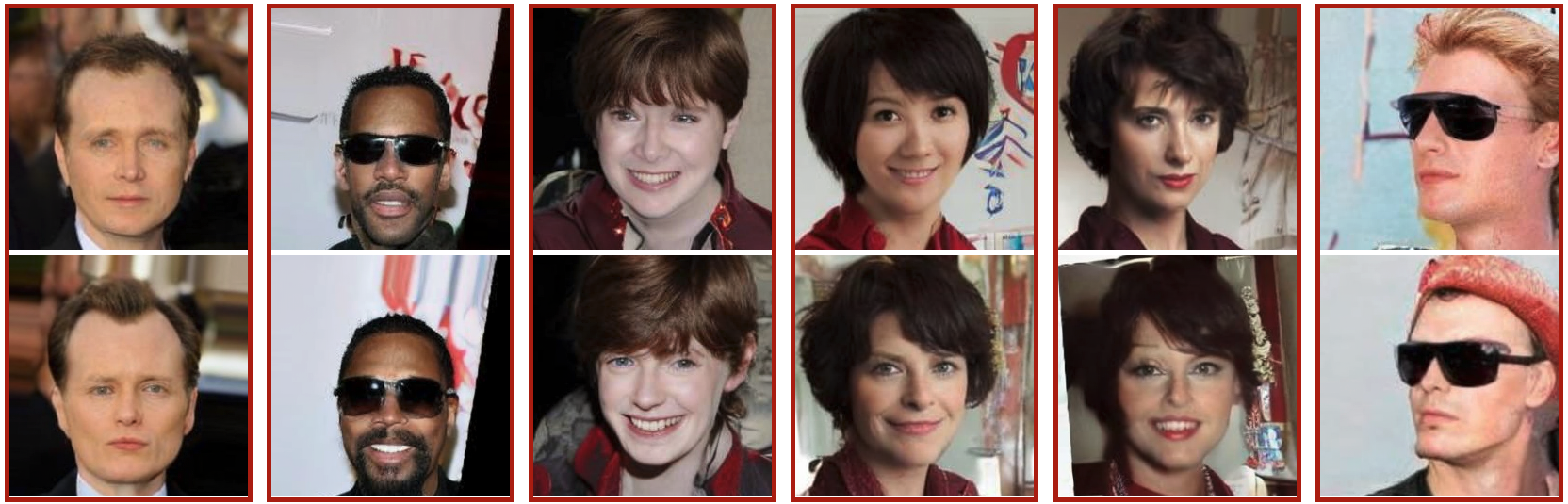}
    \vspace{-2ex}
    \caption{\small Results of scaling-controllable generation. We display six pairs of images, where each pair of images preserves the low scaling and differs in the high scaling.
    }
    \label{fig:07_steerable_generation_results}
\end{figure}

\begin{figure}[t]
    \centering
    \includegraphics[width=.8\linewidth]{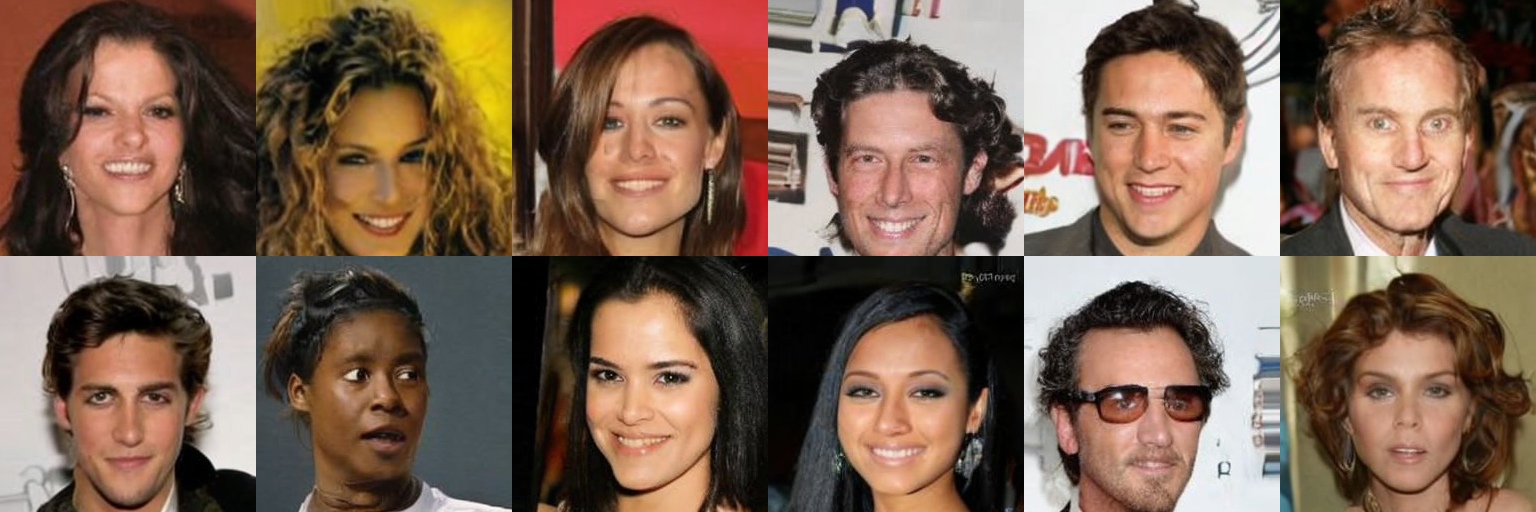}
    \vspace{-2ex}
    \caption{\small PCA editing with \model{}.}
    \label{fig:pca_editing}
\end{figure}

\textbf{Preserving Low Scaling, Modifying High Scaling.}
We need to highlight that in \model{}, when modeling the flow from lower to higher scales, the noise at higher scales is used to predict the velocity at the lower scale. This is determined by the nature of ODE flow. To this end, we conduct a study by reversing the scaling direction in the Daubechies wavelet \model{}, and the pipeline is illustrated in \Cref{fig:07_steerable_generation_pipeline}. In such a reversed setup, we keep the low-scaling part the same noise while gradually denoising the high-scaling part.\looseness=-1

The results are listed in \Cref{fig:07_steerable_generation_results}. According to the six pairs of results, we can observe that the low-scaling part stays the same, like the background of the image and the gender and color of the people, while the high-resolution details of facial expressions and outlook vary within each pair.

\textbf{Remarks.}
Although the overall results are generally optimistic, some unexpected changes have been observed in the high-scaling parts. This may be attributed to two factors: 
\begin{enumerate}[noitemsep,topsep=0pt]
    \item The compressed latent space may mix high and low content present in the original pixel space.
    \item The loss \Cref{eq: loss} may not be perfectly optimized, meaning that \model{} localized vector field might not be perfectly confined to the low-scaling part.
The second factor might be mitigated by training on larger datasets. Furthermore, by training a reversed \model{} flow (from high to low), we observe that fixing the low-scaling noise enables unsupervised editing of detailed high-scaling content.
\end{enumerate}

In \Cref{fig:07_steerable_generation_results}, we've tested the wavelet-based \model{} and observed similar results with the Fourier-based \model{}. However, for PCA, we couldn't identify obvious semantic edits that are interpretable to human eyes (see \Cref{fig:pca_editing}). This might be because PCA scaling doesn't align well with multi-resolution inductive biases.

This insight further supports our model’s capacity to decompose the generative process into distinct frequency bands, where specific frequency bands can be independently controlled. This separation aids in achieving more detailed and deliberate modifications to generated data, adding a layer of precision and flexibility to the generative framework.

\begin{table}[t]
\centering
\caption{\small
\model{} against seven generative models on COD-Cluster17 with 5K, 10K, and all samples. The best results are marked in \textbf{bold}. 
}
\label{tab:main_crystallization_result_table}
\vspace{-2ex}
\begin{adjustbox}{max width=\linewidth}
\begin{tabular}{l rr rr rr}
    \toprule
     & \multicolumn{2}{c}{\textbf{COD-Cluster17-5K}} & \multicolumn{2}{c}{\textbf{COD-Cluster17-10K}} & \multicolumn{2}{c}{\textbf{COD-Cluster17-All}} \\
     \cmidrule(lr){2-3} \cmidrule(lr){4-5} \cmidrule(lr){6-7}
     & \textbf{PM (atom) $\downarrow$} & \textbf{PM (center) $\downarrow$} & \textbf{PM (atom) $\downarrow$} & \textbf{PM (center) $\downarrow$} & \textbf{PM (atom) $\downarrow$} & \textbf{PM (center) $\downarrow$} \\
    \midrule
    GNN-MD & 13.67 $\pm$ 0.06 & 13.80 $\pm$ 0.07 & 13.83 $\pm$ 0.06 & 13.90 $\pm$ 0.05 & 22.30 $\pm$ 12.04 & 14.51 $\pm$ 0.82 \\
    CrystalSDE-VE & 15.52 $\pm$ 1.48 & 16.46 $\pm$ 0.99 & 17.25 $\pm$ 2.46 & 17.86 $\pm$ 1.11 & 17.28 $\pm$ 0.73 & 18.92 $\pm$ 0.03 \\
    CrystalSDE-VP & 18.15 $\pm$ 3.02 & 19.15 $\pm$ 4.46 & 22.20 $\pm$ 3.29 & 21.39 $\pm$ 1.50 & 18.03 $\pm$ 4.56 & 20.02 $\pm$ 3.70 \\
    CrystalFlow-VE & 14.87 $\pm$ 7.07 & 13.08 $\pm$ 4.51 & 16.41 $\pm$ 2.64 & 16.71 $\pm$ 2.35 & 12.80 $\pm$ 1.20 & 15.09 $\pm$ 0.34 \\
    CrystalFlow-VP & 15.71 $\pm$ 2.69 & 17.10 $\pm$ 1.89 & 19.39 $\pm$ 4.37 & 16.01 $\pm$ 3.13 & 13.50 $\pm$ 0.44 & 13.28 $\pm$ 0.48 \\
    CrystalFlow-LERP & 13.59 $\pm$ 0.09 & 13.26 $\pm$ 0.09 & 13.54 $\pm$ 0.03 & 13.20 $\pm$ 0.03 & 13.61 $\pm$ 0.00 & 13.28 $\pm$ 0.01 \\
    AssembleFlow & 7.27 $\pm$ 0.04 & 6.13 $\pm$ 0.10 & 7.38 $\pm$ 0.03 & 6.21 $\pm$ 0.05 & 7.37 $\pm$ 0.01 & 6.21 $\pm$ 0.01 \\
    \midrule
    \model{} (ours) & \textbf{7.21 $\pm$ 0.12} & \textbf{6.11 $\pm$ 0.11} & \textbf{7.26 $\pm$ 0.06} & \textbf{6.12 $\pm$ 0.07} & \textbf{7.23 $\pm$ 0.01} & \textbf{6.07 $\pm$ 0.01} \\
    \bottomrule
\end{tabular}
\end{adjustbox}
\end{table}

\subsection{Molecular Assembly}
We consider another scientific task: molecular assembly. The goal is to learn the trajectory on moving clusters of weakly-correlated molecular structures to the strongly-correlated structures.

\textbf{Dataset and evaluation metrics.}
We evaluate our method using the crystallization dataset COD-Cluster17~\citep{liu2024equivariant}, a curated subset of the Crystallography Open Database (COD)\citep{grazulis2009crystallography} containing 133K crystals. We consider three versions of COD-Cluster17 with 5K, 10K, and the full dataset. To assess the quality of the generated molecular assemblies, we employ \textit{Packing Matching (PM)}\citep{chisholm2005compack}, which quantifies how well the generated structures align with reference crystals in terms of spatial arrangement and packing density. Following~\citep{liu2024equivariant}, we compute PM at both the atomic level (PM-atom) and the mass-center level (PM-center)~\citep{chisholm2005compack}.

\textbf{Baselines.}
We evaluate our approach against GNN-MD~\citep{liu2024equivariant}, variations of CrystalSDE and CrystalFlow~\citep{liu2024equivariant}, and the state-of-the-art AssembleFlow~\citep{guo2025assembleflow}. CrystalSDE-VE/VP model diffusion via stochastic differential equations, while CrystalFlow-VE/VP use flow matching, with VP focusing on variance preservation. CrystalFlow-LERP employs linear interpolation for efficiency. AssembleFlow~\citep{guo2025assembleflow} enhances rigidity modeling using an inertial frame.

\textbf{Main results.}
The main results in \Cref{tab:main_crystallization_result_table} show that \model{} outperforms all baselines across three datasets. Building on AssembleFlow’s rigidity modeling, \model{} decomposes molecular pairwise distances via spectral methods and projects geometric information from $\mathbb{R}^3$ and $\text{SO}^3$ accordingly. This approach achieves consistently superior packing matching performance.

\section{Conclusion}
In this paper, we introduce K-Flow Matching (\model{}), a flow-matching model that flows along the $K$-amplitude for generative modeling. \model{} naturally defines the multi-scales of data ({\eg}, multi-resolution in image) to the multi-scales in the $K$-amplitude space.

\textbf{Future Directions on Applications.}
In this work, we verify the effectiveness of \model{} exclusively on image generation tasks. Moving forward, two promising directions are worth exploring. (1) Multi-modal generation: This includes tasks such as text-guided image generation, which could better showcase the steerability of \model{} by aligning images with natural language inputs. (2) Broader applications: Expanding the use of \model{} to a wider range of tasks, particularly those in scientific discovery, offers significant potential.\looseness=-1

\textbf{Future Directions in Theory.}
We outline six properties of \model{} in~\Cref{sec:introduction}, {\eg}, the amplitude naturally corresponds to energy. While \Cref{sec:method} briefly discusses how energy is represented in \model{}, this aspect has not been explored in depth. We believe that such energy term holds potential for integration with the utility of energy-based models in future work.\looseness=-1

\section*{Author Contribution Statement}
All authors discussed the idea. W.D. and S.L. wrote the paper and ran the experiments. F.W. and Y.R. provided the computational resources.

\section*{Acknowledgement}
The authors would like to sincerely thank Prof. Jennifer Chayes and Prof. Christian Borgs for their valuable feedback.

{
\renewcommand*{\bibfont}{\small}
\printbibliography[title={References}]
}
\clearpage
\newpage

\renewcommand{\thetable}{S\arabic{table}}
\captionsetup[table]{name=Supplementary Table}
\setcounter{table}{0}

\renewcommand{\thefigure}{S\arabic{figure}}
\captionsetup[figure]{name=Supplementary Figure}
\setcounter{figure}{0}

\addtocontents{toc}{\protect\setcounter{tocdepth}{0}}
\appendix
\section{Related Work} \label{sec: related work}

There have been multiple research lines on studying generative modeling, especially in terms of multi-scale modeling. In this work, we would like to summarize them as the following three venues.

\subsection{Multi-Scale in Pixel Resolution, Flow and Diffusion}

\textbf{Laplacian Pyramid and Laplacian Operator.} In mathematics, the Laplacian operator computes the second derivative of a function, emphasizing regions with significant intensity changes, such as edges or high-frequency details. Similarly, the Laplacian Pyramid~\citep{burt1987laplacian} decomposes an image into multiple scales, extracting the low-frequency components (smooth regions) through downsampling. The high-frequency details, such as edges and textures, are modeled as the residuals between adjacent resolution layers. The primary objective of the Laplacian Pyramid is to represent these residuals across scales in a hierarchical fashion.

\textbf{LAPGAN (Laplacian Generative Adversarial Networks)}~\citep{denton2015deep} adopts the Laplacian pyramid idea into the generative adversarial network (GAN) framework~\citep{goodfellow2014generative}. By focusing on learning residuals between successive levels of resolution, it effectively generates high-quality super-resolution images.

\textbf{SR3 (Super-Resolution via Repeated Refinement)}~\citep{saharia2022image} leverages DDPM (Denoising Diffusion Probabilistic Models)~\citep{ho2020denoising} and DSM (Denoising Score Matching)~\citep{vincent2011connection,song2019generative} for high-resolution image generation. Specifically, SR3 enhances low-resolution images to high-resolution by utilizing multiple cascaded conditional diffusion models. In this framework, the low-resolution images serve as conditions, and the model's aim is to predict the corresponding high-resolution images as outputs.

\textbf{PDDPM (Pyramidal Denoising Diffusion Probabilistic Models)}~\citep{ryu2022pyramidal} is a follow-up work of SR3, and it improves the model by only modeling one score network. The key attribute to enable this is by adding the fractional position of each pixel to the score network, and such fractional position information can be naturally generalized to different resolutions.

\textbf{f-DM}~\citep{gu2022f} is developed concurrently with PDDPM and shares the approach of utilizing only one diffusion model. It distinguishes itself by explicitly applying a sequence of transformations to the data and emphasizing a resolution-agnostic signal-to-noise ratio within its diffusion model design.

\textbf{Edify Image}~\citep{atzmon2024edify} is a state-of-the-art model capable of generating photorealistic, high-resolution images from textual prompts \citep{atzmon2024edify}. It operates as a cascaded pixel-space diffusion model. To enhance its functionality, Edify Image employs a downsampling process that extracts low-frequency components and creates three distinct resolution levels, ranging from low to high frequency, with the original image representing the highest frequency level. Another key innovation of Edify Image is its meticulously crafted training and sampling strategies at different resolutions, utilizing attenuated noise schedules.

\subsection{Multi-Scale in Pixel Resolution, VAE and AR}

\textbf{VQ-VAE2 (Vector Quantized VAE 2)}~\citep{razavi2019generating} enforces a two-layer hierarchical structure, where the top layer captures global features such as object shapes and geometry, while the bottom layer focuses on local details like texture. It models data density within the variational autoencoder (VAE) framework\citep{kingma2013auto} and incorporates an autoregressive (AR) module to enhance the prior for improved generative performance.

\textbf{RQ-VAE (Residual-Quantized VAE)}~\citep{lee2022autoregressive} integrates recursive quantization into the VAE framework. It constructs a representation by aggregating information across $D$ layers, where the first layer provides a code embedding closely aligned with the encoded representation, and each subsequent layer refines this by reducing the quantization error from the previous layer. By stacking $D$ layers, the accumulated quantization error is minimized, enabling RQ-VAE to offer a coarse-to-fine-grained approach to modeling. For modeling, the general pipeline follows the VAE framework, while each latent code is decomposed into $D$ layers and is predicted in an autoregressive manner.

\textbf{VAR (Visual AutoRegressive)}~\citep{tian2024visual} introduces a novel paradigm for density estimation by decomposing images into multiple resolutions across various scales. This approach is inspired by the hierarchical nature of human perception, where images are interpreted progressively from global structures to finer details. Leveraging this concept, VAR models the entire image in a coarse-to-fine manner, adhering to the principles of multi-scale hierarchical representation.

\subsection{Multi-Scale in Frequency, AR and Diffusion}

\textbf{WaveDiff (Wavelet Diffusion)}~\citep{phung2023wavelet} leverages the discrete wavelet transform to shift the entire diffusion process into the wavelet spectrum. Its primary objective is to reduce model complexity by operating in the transformed spectrum space instead of the pixel domain.

\textbf{PiToMe (Protect Informative Tokens before Merging)}~\citep{tran2024accelerating} is a token merging method designed to balance efficiency and information retention. PiToMe identifies large clusters of similar tokens as high-energy regions, making them suitable candidates for merging, while smaller, more unique, and isolated clusters are treated as low-energy and preserved. By interpreting attention over sequences as a fully connected graph of tokens, PiToMe leverages spectral graph theory to demonstrate its ability to preserve critical information.

\textbf{SIT (Spectral Image Tokenizer)}~\citep{esteves2024spectral} is a parallel work to ours that processes the spectral coefficients of input patches (image tokens) obtained through a discrete wavelet transform. Motivated by the spectral properties of natural images, SIT focuses on effectively capturing the high-frequency components of images. Furthermore, it introduces a scale-wise attention mechanism, referred to as scale-causal self-attention, which is designed to improve the model's expressiveness across multiple scales.

\end{document}